\documentclass{article}

\PassOptionsToPackage{comma, numbers, sort}{natbib}

\usepackage[final]{neurips_2025}

\usepackage[utf8]{inputenc} %
\usepackage[T1]{fontenc}    %
\usepackage{hyperref}       %
\usepackage{url}            %
\usepackage{booktabs}       %
\usepackage{amsfonts}       %
\usepackage{nicefrac}       %
\usepackage{microtype}      %
\usepackage{xcolor}         %
\usepackage{graphicx}
\usepackage{amsmath}
\usepackage{subcaption}
\usepackage{float}
\usepackage{algorithm}
\usepackage{algorithmic}
\usepackage[frozencache=true, cachedir=minted-cache]{minted}
\usepackage[capitalise,noabbrev,nameinlink]{cleveref}
\DeclareMathOperator*{\argmin}{arg\,min}

\newcommand{\maincolor}[1]{\textcolor[HTML]{d21418}{#1}}

\definecolor{mainc}{HTML}{d21418}
\hypersetup{colorlinks=true, urlcolor=mainc,citecolor=mainc,linkcolor=mainc}

\title{A Stable Whitening Optimizer for \\ Efficient Neural Network Training}

\author{%
  Kevin Frans \\
  UC Berkeley \\
  \texttt{kvfrans@berkeley.edu} \\
  \And
  Sergey Levine \\
  UC Berkeley \\
  \And
  Pieter Abbeel \\
  UC Berkeley \\
}

\begin{document}

\maketitle

\begin{abstract}
    In this work, we take an experimentally grounded look at neural network optimization. Building on the Shampoo family of algorithms, we identify and alleviate three key issues, resulting in the proposed SPlus method. First, we find that naive Shampoo is prone to divergence when matrix-inverses are cached for long periods. We introduce an alternate bounded update combining a historical eigenbasis with instantaneous normalization, resulting in across-the-board stability and significantly lower computational requirements. Second, we adapt a shape-aware scaling to enable learning rate transfer across network width. Third, we find that high learning rates result in large parameter noise, and propose a simple iterate-averaging scheme which unblocks faster learning. To properly confirm these findings, we introduce a pointed Transformer training benchmark, considering three objectives (language modelling, image classification, and diffusion modelling) across different stages of training. On average, SPlus is able to reach the validation performance of Adam within \maincolor{$44-58\%$} of the gradient steps and \maincolor{$62-83 \%$} of the wallclock time.
\end{abstract}

\section{Introduction}

The backbone of modern deep learning is stochastic gradient descent -- a method which, while particularly effective in practice, is time-consuming by its iterative nature. As large neural networks are scaled up to billions of parameters \citep{brown2020language, podell2023sdxl, achiam2023gpt}, and are trained on datasets of similar scale, it becomes increasingly important that each gradient update is efficient in making learning progress. 

A core optimization strategy involves adapting updates to second-order curvature \citep{lecun2002efficient, amari1998natural}, allowing for faster learning progress in certain directions while preventing divergence in others. However, practical challenges arise with regards to large neural network training. For example, it is computationally intractable to calculate the Hessian of any reasonably-sized network \citep{martens2010deep, li2017preconditioned}, and even further intractable to invert this matrix. Numerical stability is also a concern, and complex methods often require additional hyperparameter tuning and regularization \citep{martens2015optimizing, agarwal2020disentangling}.

In this work, we take a cautiously empirical view on neural network optimization. We carefully design an evaluation suite that is well-aligned with common uses today. We consider the optimization of a standard Transformer model \citep{vaswani2017attention} on \textit{three distinct training settings} -- autoregressive language modelling \citep{radford2019language}, diffusion modelling \citep{ho2020denoising, peebles2023scalable}, and image classification \citep{dosovitskiy2020image}. We further consider the performance of optimizers at various stages of training (e.g. starting from checkpoints near the start, middle, and end), to avoid biasing towards early-stage performance gains. To our knowledge, this is currently the widest-scope comparison of adaptive optimizers on Transformer training.

From findings in this setting, we develop a method we refer to as SPlus. SPlus builds on the Shampoo family of algorithms \citep{gupta2018shampoo}, which can broadly be seen as approximating a \textit{whitening metric} from historical gradients, and performing steepest descent along this metric. We identify and address three issues with the naive Shampoo update. First, we find that Shampoo often diverges at high learning rates or under less-frequent inversion rates. To fix this issue, we introduce an alternative bounded update combining the historical eigenbasis with instantaneous normalization, resulting in across-the-board stability even with significantly lower-frequency inversions. Second, we find that Shampoo update magnitudes are not properly scaled with relation to network width. We adopt methodology derived for SGD and Adam \citep{yang2023spectral} to our setting, enabling easy hyperparameter tuning via learning rate transfer across network widths. Third, we find that higher learning rates in whitening-based optimizers result in significant parameter noise. We propose a simple iterate-averaging scheme to alleviate this issue, unblocking fast learning with a much lower degradation in performance.

In experiments, SPlus achieves the fastest loss decrease among a wide suite of previous optimizers. SPlus matches Adam within $\mathbf{\maincolor{\sim 44-58\%}}$ of the gradient steps, and within $\mathbf{\maincolor{\sim 62-83 \%}}$ of the wallclock time. SPlus does not introduce any critical hyperparameters to tune, and does not require additional forward/backward passes. We hope that this effectiveness and simplicity enables the community to easily adopt SPlus to their existing training settings.

We provide the full code to replicate experiments at \href{https://github.com/kvfrans/splus}{github.com/kvfrans/splus}. The repo also contains single-file implementations of SPlus in JAX and Pytorch, along with basic reccomendations for usage.

\vspace{-4pt}
\section{Related Work}
\vspace{-4pt}
\textbf{Second-order optimization of neural networks.} Second-order methods, which we broadly define as methods which further modify the first-order gradient, can be largely categorized among two axes -- the second-order metric used, and the way in which the metric is approximated and applied. A common metric choice is the Hessian, which in Newton's method can be directly applied by multiplying the inverse Hessian and the gradient. Neural-network specific methods have approximated the Hessian via conjugate-gradient methods \citep{martens2010deep}, iterative fitting \citep{li2017preconditioned, pooladzandi2024curvature}, or a diagonal approximation \citep{lecun2002efficient, liu2023sophia}. An alternate metric is the Fisher, motivated by natural gradient methods \citep{amari1998natural, sohl2012natural}. Closest related to our work is the use of the "empirical" Fisher metric \citep{kunstner2019limitations}, i.e. an uncentered covariance matrix of gradients. The most well-tested of these are the diagonal Adam and its variants \citep{duchi2011adaptive, kingma2014adam, loshchilov2017decoupled, shazeer2018adafactor}.

\textbf{Kronecker factorized metrics.} Our work utilizes a Kronecker factorization of the empirical Fisher, building on the techniques introduced in K-FAC \citep{martens2015optimizing}, and more directly Shampoo \citep{gupta2018shampoo}. Shampoo has been extended to utilize distributed inversion and exponential averaging \citep{shi2023distributed}, and/or perform inversion via Newton iteration \citep{anil2020scalable}. Closest related to our work is SOAP \citep{vyas2024soap}, which similarly utilizes an eigen-decomposition of the factor matrices, however, SOAP maintains an additional set of second-moments per parameter, whereas in our work this is not neccessary.

\textbf{Non-Euclidean gradient descent.} We build on previous works which have studied gradient descent over non-Euclidean distance metrics. In particular, such a framing may manifest not as linear transformations of the gradient, but also methods which explicitly parameterize learning rate \citep{yang2023spectral, large2024scalable}, or instantaneously transform the gradient via sign descent \citep{bernstein2024old}. Newton-Shulz orthogonalization of gradients \citep{jordanmuon, ma2024swan} can also be motivated as maximum spectral descent.

\textbf{Iterate averaging.} Whitening-based optimizers have poor convergence properties due to the uniform size of each update, and instead typically rely on learning-rate decay. An alternative strategy is to average fast-moving snapshots of parameters \citep{ruppert1988efficient, polyak1990new, polyak1992acceleration, izmailov2018averaging, morales2024exponential}, which has been shown effective in approximating learning-rate decay \citep{defazio2024road} and is often used in image generation models \citep{yaz2018unusual, karras2024analyzing}.
\vspace{-4pt}
\section{Background and Preliminaries}
\vspace{-4pt}

Gradient descent methods can be seen as following the steepest descent direction under some metric. Naive gradient descent implicitly assumes a Euclidean distance metric over parameters, in which case the update vector is simply the scaled gradient $g = \nabla_\theta L(\theta, x)$ itself:
\begin{equation}
    u =\argmin_{\Delta\theta} \; \underbrace{\; g^T\Delta\theta \;}_{\text{Improvement}} + \underbrace{(\alpha/2)||\Delta\theta||^2}_{\text{Distance Penalty}} \quad = \quad \alpha g.
    \label{eq:naivegd}
\end{equation}
However, it is often helpful to impose other metrics. For example, certain parameters may be more sensitive to second-order changes, and thus should be assigned a larger penalty. We can generally express distance using a Riemannian metric tensor $M$, which is a symmetric positive-definite matrix of shape $\mathbb{R}^{dim(\theta) \times dim(\theta)}$.
Under $M$, the distance of an update can be expressed as the matrix product:
\begin{equation}
    ||\Delta\theta||^2_M = \Delta\theta^T M \Delta\theta.
\end{equation}
Gradient descent can now be performed using $M$ as the distance metric. The solution then becomes:
\begin{equation}
    u = \argmin_{\Delta\theta} \; \underbrace{\; g^T\Delta\theta \;}_{\text{Improvement}} + \underbrace{(1/2)\Delta\theta^TM\Delta\theta}_{\text{Distance Penalty}} \quad = \quad M^{-1}g.
\end{equation}
which, in the case of Euclidean distance represented by the identity metric, reduces to \Cref{eq:naivegd}.

\begin{figure}
    \centering
    \includegraphics[width=\textwidth]{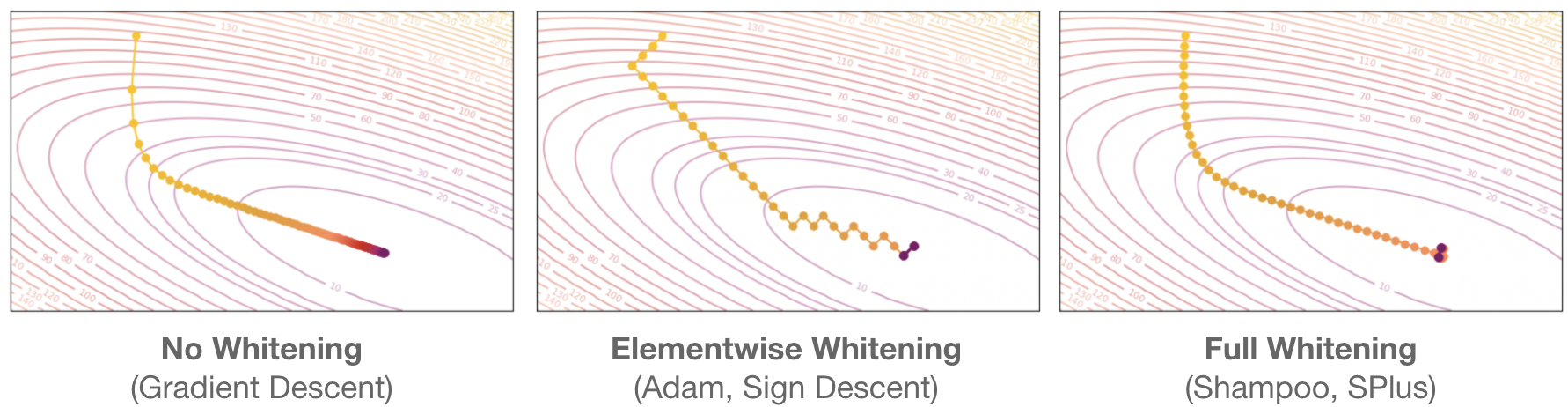}
    \caption{\textbf{Whitening normalizes gradients to have uniform magnitude along each axis of descent.} This decouples the updates from gradient magnitude. Elementwise whitening imposes an independent axis per dimension, whereas full whitening uses the axes that maximally explain gradient covariance.}
    \label{fig:whitening}
\end{figure}

\textbf{Whitening metric.} Empirically, the \textit{whitening metric} \citep{yang2008principal} has proven to be a reliable choice for neural network optimization. The whitening metric is the square-root of the uncentered covariance:
\begin{equation}
    M = \mathbb{E}_x \left[ \nabla_\theta L(\theta,x) \nabla_\theta L(\theta,x)^T \right]^{1/2} = \mathbb{E}_x \left[ gg^T \right]^{1/2}
\end{equation}
and is guaranteed to be positive definite. The whitening metric can be calculated via empirical gradients and does not require additional forward/backward passes. Notably, Adam \citep{kingma2014adam} performs descent on a diagonal approximation of the whitening metric, assuming each parameter as independent.

\textbf{What about the Hessian or the Fisher?} While potentially effective, these metrics are often expensive to compute as they require additional information outside the standard gradient. We refer to detailed discussions in \citep{mccandlish2018empirical, martens2020new, sohl2012natural, korbit2024exact}, as well as a brief overview in \cref{sec:preconditioning}.

\textbf{Approximations for neural network learning.} To make storing and inverting the metric amenable for large neural networks, it is common to assume a per-layer blockwise approximation \citep{martens2015optimizing, gupta2018shampoo}. In this way, the full whitening metric can be represented as a set of smaller block matrices, one per layer, which can each be independently inverted. 

To further reduce memory and computation, each block can be further approximated by a Kronecker product of two smaller matrices:
\begin{equation}
     M^{mn,mn} =A^{m,m} \otimes B^{n,n} =
\begin{bmatrix}
a_{11}B & a_{12}B & \cdots & a_{1m}B \\
\vdots & \vdots & \ddots & \vdots \\
a_{m1}B & a_{m2}B & \cdots & a_{mm}B
\end{bmatrix}.
\end{equation}
Kronecker products have a useful property that the inverse (at any power) of a Kronecker product is equivalent to the Kronecker product of the inverse factors:
\begin{equation}
    \text{if} \qquad M = A \otimes B, \quad \text{then} \quad M^{-1/2} = (A^{-1/2} \otimes B^{-1/2}).
\end{equation}
Additionally, multiplication by a Kronecker product can be performed without explicitly forming the full product matrix. Consider the flattened gradient and update vectors $g,u \in \mathbb{R}^{mn}$ and their corresponding matrix forms $G,U \in \mathbb{R}^{m \times n}$. The following operations are identical:
\begin{equation}
    u= (A \otimes B)^{-1/2}g \quad \leftrightarrow \quad U = A^{-1/2} GB^{-1/2}
\end{equation}
In Shampoo \citep{gupta2018shampoo} (\cref{alg:shampoo}), the above techniques are utilized to derive an efficient update. The factor matrices of the whitening metric can be directly calculated from the matrix-shaped gradients:
\begin{equation}
    M = E[gg^T]^{-1/2} \quad \approx \quad (L \otimes R) \quad \text{where} \quad L = E[GG^T]^{-1/4} \qquad R = E[G^TG]^{-1/4}
\end{equation}
after which the update can be calculated as:
\begin{equation}
    u = M^{-1}g \quad \approx \quad U = L^{-1/4} G R^{-1/4}.
\end{equation}

In practice, matrix inversion is slow, so the above inversion is performed only every $N$ steps, and the results are cached until recomputation. This caching can result in unstable training (\cref{fig:signdescent}), and is one of the issues we will discuss in the following sections.

\section{SPlus: A Stable Whitening Optimizer}

Our main contribution is SPlus, an efficient optimizer which builds upon Shampoo to stabilize training and reduce overall gradient-step and wall-clock time. In developing SPlus, we take a fundamentally empirical and experimental approach -- we identify three core shortcomings of the naive Shampoo method, examine their causes, and propose a series of nuanced improvements to alleviate these issues. In aggregate, these changes lead to a significant improvement in reliability and training speed.

\subsection{To reliably prevent divergence, utilize instant-sign normalization}

\begin{figure}
    \centering
    \includegraphics[width=0.9\textwidth]{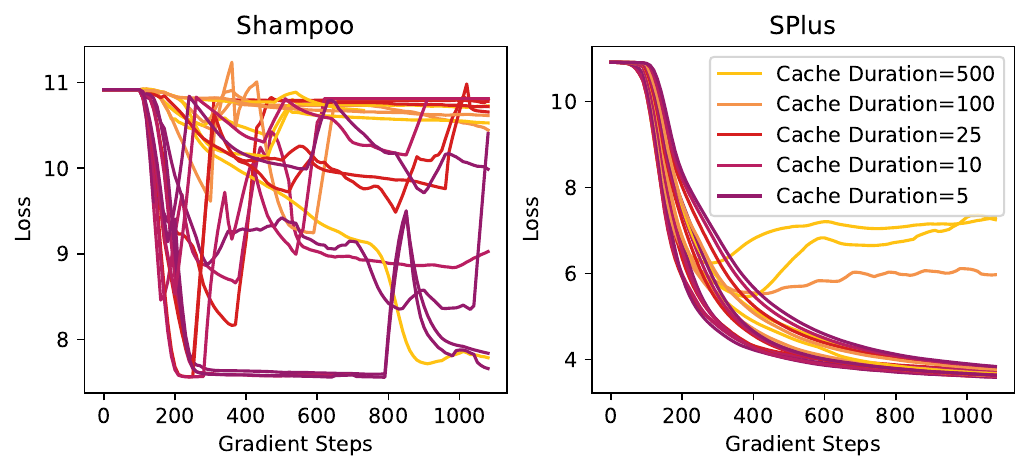}
    \caption{\textbf{Shampoo is prone to divergence, but SPlus remains stable under the same settings. \newline}  Plotted above are loss curves on language modelling, sweeping over learning rate between $(0.0001, 0.000215, 0.000464, 0.001)$ and cache duration between $(5,10,25,100,500)$. SPlus is significantly more robust to hyperparameters than Shampoo. \textit{This robustness is crucial for improving practical training speed} -- in our setting. Shampoo diverges when caching for $>100$ gradient steps while SPlus remains stable, enabling a faster wall-clock performance than Adam.}
    \label{fig:signdescent}
\end{figure}

As shown in \cref{fig:signdescent}, we find that naive Shampoo is prone to divergence. When examining a range of learning rates and matrix-inversion frequencies, we find that Shampoo diverges in $> 50\%$ of the trials in our setting. Notably, Shampoo regularly diverges when matrix-inversion is cached for over 25 gradient steps.

We hypothesize that the interaction between \textit{cached} matrix-inverses and incoming gradients is the cause of frequent divergence. To provide intuition on this behavior, we can rewrite the square-root matrix inverse in terms of its eigen-decomposition\footnote{Eigen-decomposition is commonly used under the hood for calculating symmetric matrix inverses, as matrix powers share the same eigenbasis, and the diagonal term is easily raised to any power.}:
\begin{equation}
    U_{Shampoo} = (Q_L \Lambda_L^{-1/4} Q_L^T) \; G \; (Q_R \Lambda_R^{-1/4} Q_R^T)
\end{equation}
where eigenvectors $Q$ are orthonormal, and eigenvalues $\Lambda$ are diagonal. 

In the above decomposition, the eigenvectors can be understood as basis directions that maximally explain the covariance between gradients. Each eigen\textit{value} represents the historical squared magnitude of gradients along each basis. The Shampoo update normalizes incoming gradients by their respective historical magnitudes along each basis.

The risk is when incoming gradients align with a tail-end basis direction (which has a small historical magnitude), in which case the update can diverge. This risk is especially prominent when the cached matrix-inverse is stale, as incoming gradients may no longer align with the historical distribution.

To alleviate this risk, we instead propose a normalization scheme that does not rely on historical magnitudes at all. Sign-normalization has been previously studied as the equivalence of Adam without a running average, accomplishing a similar normalizing behavior \citep{bernstein2018signsgd, bernstein2024old}. We therefore opt to ignore historical magnitudes, and instead perform normalization instantaneously via the 'sign' function. The eigenbasis remains as the historical eigenbasis. We refer to this update as \textit{instant-sign normalization}:
\begin{equation}
    U = Q_L sign(Q_L^T G Q_R) Q_R^T.
\end{equation}
Instant-sign normalization has a hard bound preventing divergence. As $Q_L/Q_R$ are orthonormal, and the inner sign-matrix contains only $1$ or $-1$, the resulting update will satisfy:
\begin{equation}
    ||U||_{spectral} \leq ||U||_{frob} = \sqrt{nm} \qquad \text{and} \qquad ||U||_\infty \leq \max(\sqrt{m}, \sqrt{n}).
    \label{eq:norm}
\end{equation}

Additionally, instant-sign normalization provides a more fine-grained elementwise normalization than naively using the Kronecker-approximated inverse factors. As a motivating example, under an identity eigenbasis, the Shampoo update would be:
\begin{equation}
    U = \Lambda_L^{-1/4} G \Lambda_R^{-1/4}  \quad \leftrightarrow \quad u =(\lambda_L \otimes \lambda_R)^{-1/4} \; g
\end{equation}
where notably, the diagonal component of $(\lambda_L \otimes \lambda_R)$ is not fully expressive due to being constructed out of a Kronecker product. This notion is studied in SOAP \citep{vyas2024soap}, who note that Shampoo (with a $1/2$ power) is equivalent to a rank-1 Adam approximation in a rotated eigenbasis. Their proposed method alleviates this issue with an additional elementwise normalization matrix for each parameter. In contrast, instant-sign normalization does not require additional parameters in memory. 

As shown in \cref{fig:signdescent} (left), the instant-sign normalization of SPlus eliminates divergence across the board. Crucially, SPlus allows for the matrix-inversion to be cached for significantly longer intervals without collapse. Prior works on Shampoo perform recomputation every 10 steps \citep{shi2023distributed, gupta2018shampoo}, and our empirical findings support that this frequency is needed to prevent divergence. In contrast, SPlus remains stable even when results are cached for over 100 steps. Utilizing this stability, SPlus can be run at a speed which outperforms Adam in reaching an equivalent validation loss not only in gradient steps, but also in \textit{wall-clock time} (\cref{fig:signdescent}, right).

\subsection{To standardize learning rate across network widths, use symmetric shape-aware scaling}
\label{sec:scaling}

\begin{figure}
    \centering
    \includegraphics[width=\textwidth]{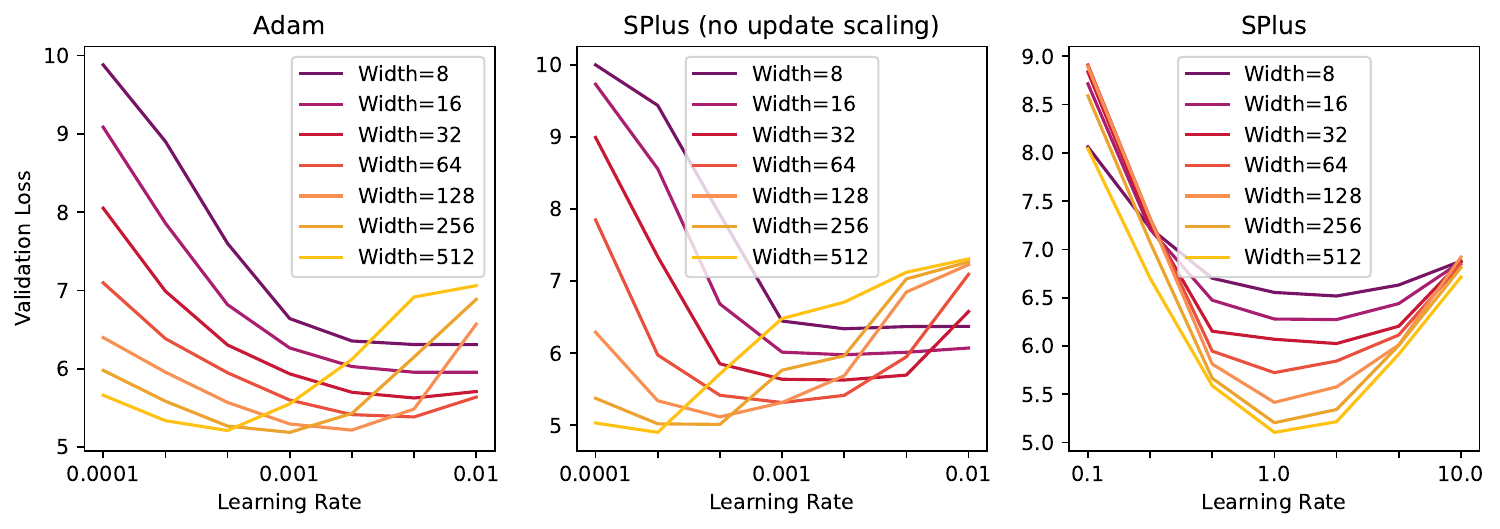}
    \caption{\textbf{Optimal learning rates for SPlus transfer across network widths.} This is achieved by normalizing per-layer update magnitudes by constant shape-dependent factor. Notably, this learning rate transfer does not hold by default for Adam or Shampoo.}
    \label{fig:transfer}
\end{figure}

Learning rate is often the first hyperparameter to tune due to its outsized impact on performance. Recent works have shown that for SGD and Adam, it is possible to naturally parameterize updates such that the optimal learning rate remains constant even as network width is adjusted \citep{yang2022tensor, yang2023spectral}. As shown in \cref{fig:transfer}, neither Shampoo nor the instant-sign update above display the correct learning-rate transfer across widths. In this section, we derive a simple adjustment to introduce learning-rate transfer to our setting as well.

We start by defining a desired property, following \citep{yang2023spectral} -- after an update, the expected magnitude of change in \textit{individual intermediate activations} should be invariant, regardless of network width:
\begin{equation}
    \text{for any intermediate activation vector $x$:} \quad \sqrt{\dfrac{1}{k}\sum_{i=0}^k (\Delta x_i)^2} = \mathbb{O}(1).
\end{equation}
To achieve the above property, the norm of the update to each dense layer of shape $U =\mathbb{R}^{m \times n}$ must be properly scaled. In SPlus, \cref{eq:norm} states that the Frobenius norm of a raw instant-sign update is $\sqrt{nm}$. Thus, when considering a network with $c$-times larger width, one should divide the update by a factor of $c$. We implement this scaling without a reference width by introducing a per-layer scaling factor of $2/ (m+n)$:
\begin{equation}
    U = Q_L sign(Q_L^T G Q_R) Q_R^T \; * \; 2 \;/\; (m+n)
    \label{eq:scaling}
\end{equation}
As shown in \cref{fig:transfer}, the shape-aware scaling factor enables a natural parametrization where learning rates transfer between network width. This property is especially useful for tuning, and enables a more robust default learning rate for the optimizer.

Our scaling factor is different than the "spectral" scaling of $1/m$ introduced in \citep{yang2022tensor}, and this is by design. We found that spectral scaling was harmful in the MLP block. Consider a $(256, 1024)$ layer and a $(1024, 256)$ layer. Under spectral scaling, the first layer would have a 4x larger per-parameter learning rate than the second. In contrast, our method is symmetric and assigns the same learning rate to both layers (as Adam does) while properly normalizing when both widths are increased. We made this design choice purely on empirical findings -- our average scaling outperformed the spectral scaling in all cases, see \cref{fig:scaling} -- and provide a brief discussion in \cref{section:symmetric}.

\subsection{To reduce parameter noise, make use of iterate averaging}

\begin{figure}
    \centering
    \includegraphics[width=\textwidth]{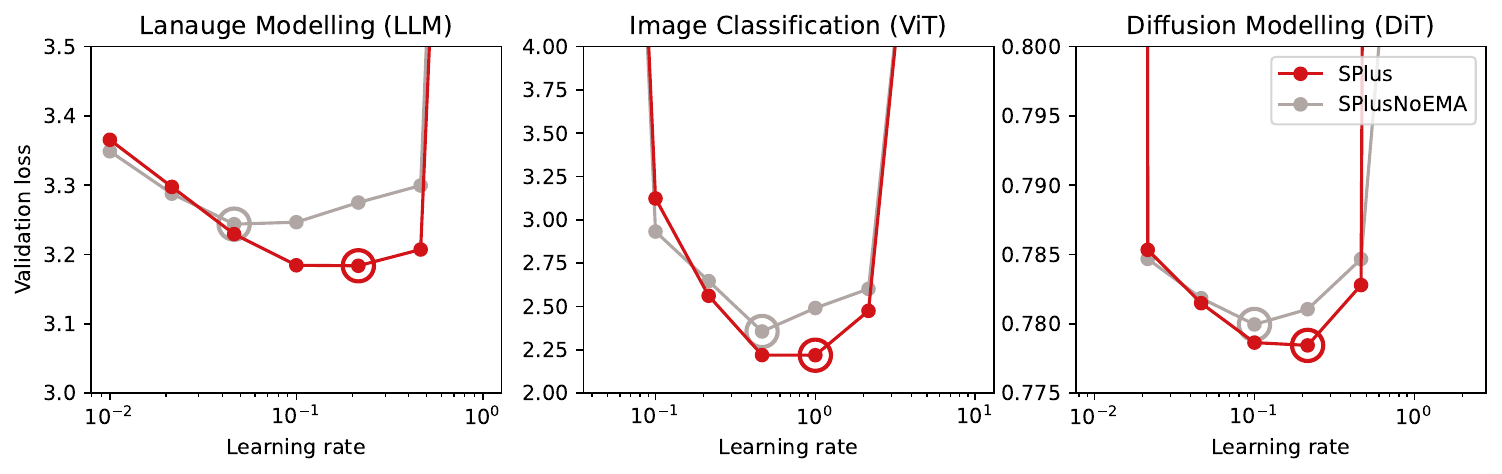}
    \caption{\textbf{Iterate averaging enables the use of higher learning rates without degradation.} Training with a higher learning rate creates a tradeoff between faster learning progress and increased parameter noise. By averaging previous iterates, parameter noise is lessened, and we can surpass the tradeoff to reveal a stronger optimal learning rate.}
    \label{fig:ema}
\end{figure}

Stochastic gradient  descent  methods inherently follow noisy descent directions. This noise can be broadly categorized as 1) noise from using a stochastically sub-sampled batch rather than the full dataset \citep{smith2017don, mccandlish2018empirical}, and 2) noise from linearizing the loss and taking a discrete step. 

The second noise source is especially prominent in whitening-based optimizers. In naive gradient descent, the magnitude of updates will decrease as the magnitude of gradients decreases, providing a natural annealing. However, whitening-based optimizers instead utilize \textit{normalized} updates which aim to take uniformly-sized updates. This behavior manifests in trajectories where parameters ``overshoot" their ideal value, and oscillate back and forth.

While learning rate decay (and lower learning rates in general) can address these issues, they result in a tradeoff of slower learning progress. At low learning rates, learning progress is slow, and loss magnitude remains large. At high learning rates, loss is again large, but for a different reason -- the presence of large noise in the parameters. Learning rate tuning can be used to locate a tradeoff between these two factors, however, is there is a way to get the best of both worlds?

We find that a more effective method of controlling the noise-progress tradeoff is via iterate averaging \citep{polyak1990new, polyak1992acceleration, ruppert1988efficient}. Specifically, a set of live parameters are updated with a large learning rate. A second set of slow parameters are calculated via an exponential moving average of the live parameters:
\begin{equation}
    \theta' \leftarrow \nabla_{\theta'} L(\theta') \qquad \theta \leftarrow (1-\beta) \theta  + \beta \theta'.
\end{equation}
In this way, learning progress can remain fast, yet the effect of gradient noise is diminished, as discussed in \cite{izmailov2018averaging} and \cite{morales2024exponential}. We note that parameter averaging is a common technique in machine learning and has been effective in a range of domain-specific methods, e.g. image generation \citep{karras2024analyzing, yaz2018unusual}, reinforcement learning \citep{van2016deep, fujimoto2018addressing}, and representation learning \citep{grill2020bootstrap}.

\cref{fig:ema} highlights the benefits of simple iterate averaging. Across the board, evaluating at the exponentially averaged parameters achieves a lower validation loss. Note that series of live parameters in the averaged and non-averaged cases are equivalent. The averaged parameters more closely reveal the ``true'' learning progress of utilizing a higher learning rate, which is otherwise obscured by parameter noise causing an increase in validation loss.

\begin{figure}
\vspace*{-4ex}
\begin{minipage}[t]{0.5\textwidth}
\begin{algorithm}[H]
\caption{Shampoo}
\begin{algorithmic}
\FOR{ \textbf{each layer gradient} $G$}
    \STATE $G = \nabla_\theta L(\theta, x)$ where $G \in \mathbb{R}^{m \times n}$
    \STATE $L \leftarrow (1-\beta_2)L+\beta GG^T$
    \STATE $R \leftarrow (1-\beta_2)R+\beta G^TG$
    \STATE $\bar{G} \leftarrow (1-\beta_1) \bar{G}  + \beta G $
    \IF{$n \bmod N = 0$}
        \STATE $\tilde{L}^{-1/4} \leftarrow matpow(L, -1/4)$
        \STATE $\tilde{R}^{-1/4} \leftarrow marpow(R, -1/4)$
    \ENDIF
    \STATE $ U \leftarrow \tilde{L}^{-1/4} \bar{G} \tilde{R}^{-1/4}$
    \vspace{0.5ex}
    \STATE $\theta \leftarrow \theta + \alpha U $
\ENDFOR
\end{algorithmic}
\label{alg:shampoo}
\end{algorithm}
\end{minipage}
\hfill
\begin{minipage}[t]{0.5\textwidth}
\begin{algorithm}[H]
   \caption{SPlus \maincolor{(changes in red)}}
\begin{algorithmic}
\FOR{ \textbf{each layer gradient} $G$}
    \STATE $G = \nabla_\theta L(\maincolor{ \theta'}), x)$ where $G \in \mathbb{R}^{m \times n}$
    \STATE $L \leftarrow (1-\beta_2)L+\beta GG^T$
    \STATE $R \leftarrow (1-\beta_2)R+\beta G^TG$
    \STATE $\bar{G} \leftarrow (1-\beta_1) \bar{G}  + \beta G $
    \IF{$n \bmod N = 0$}
        \STATE \maincolor{$Q_L, \Lambda_L \leftarrow eigh(L)$}
        \STATE \maincolor{$Q_R, \Lambda_R \leftarrow eigh(R)$}
    \ENDIF
    \STATE \maincolor{$ U \leftarrow Q_L sign(Q_L^T \bar{G} Q_R) Q_R^T * 2/(m+n)$}
    \vspace{0.5ex}
    \STATE $\theta' \leftarrow \theta' + \alpha U $
    \STATE \maincolor{$\theta \leftarrow (1-\beta_3) \theta  + \beta \theta' $}
\ENDFOR
\end{algorithmic}
\label{alg:splus}
\end{algorithm}
\end{minipage}
\end{figure}

\section{How does SPlus compare to prior optimizers?}
\label{sec:comparison}

We now present a thorough evaluation of SPlus alongside previous optimizers. Intentionally, we focus specifically on the Transformer architecture \citep{vaswani2017attention}, as it has been adapted as the backbone for most large-scale neural networks today, regardless of domain or modality \citep{peebles2023scalable, dosovitskiy2020image, team2024octo, ma2024latte}. Transformers are a general architecture. This flexibility means that we must be careful in evaluating their training, to avoid overfitting on a specific domain, e.g. only language modelling. To our knowledge, our setting currently represents the widest-scope evaluation of optimizers on Transformer training.

To demonstrate robustness across settings, we examine neural networks trained on three different objectives and datasets. First, we examine an autoregressive language model \textbf{(LLM)}, trained on the OpenWebText \citep{Gokaslan2019OpenWeb} dataset with a sequence length of 256. Second, we examine a latent diffusion model \textbf{(DiT)} \citep{peebles2023scalable}, trained via flow-matching \citep{lipman2022flow} to generate Imagenet images encoded via a pretrained variational auto-encoder \citep{rombach2022high}. Third, we examine an image classification network \textbf{(ViT)} \citep{dosovitskiy2020image}, trained to classify raw-pixel Imagenet images. All three settings are adapted directly from prior work, and utilize the same Transformer backbone.

The specific Transformer architecture is adapted from GPT-2 \citep{brown2020language}. Layer normalization terms are applied pre-attention and pre-MLP. We remove bias terms from the network. Each objective also includes different input/output heads -- a token embedding and logit predictor for language modelling, a patch embedder and patch output for diffusion modelling, and a patch embedding and class predictor for image classification. We use a momentum of $0.9$ when applicable, a linear warmup of 200 steps followed by a constant schedule, and a weight decay of $0.1$. We train in bfloat16. See the provided code for further details.

\begin{figure}
    \centering
    \includegraphics[width=\textwidth]{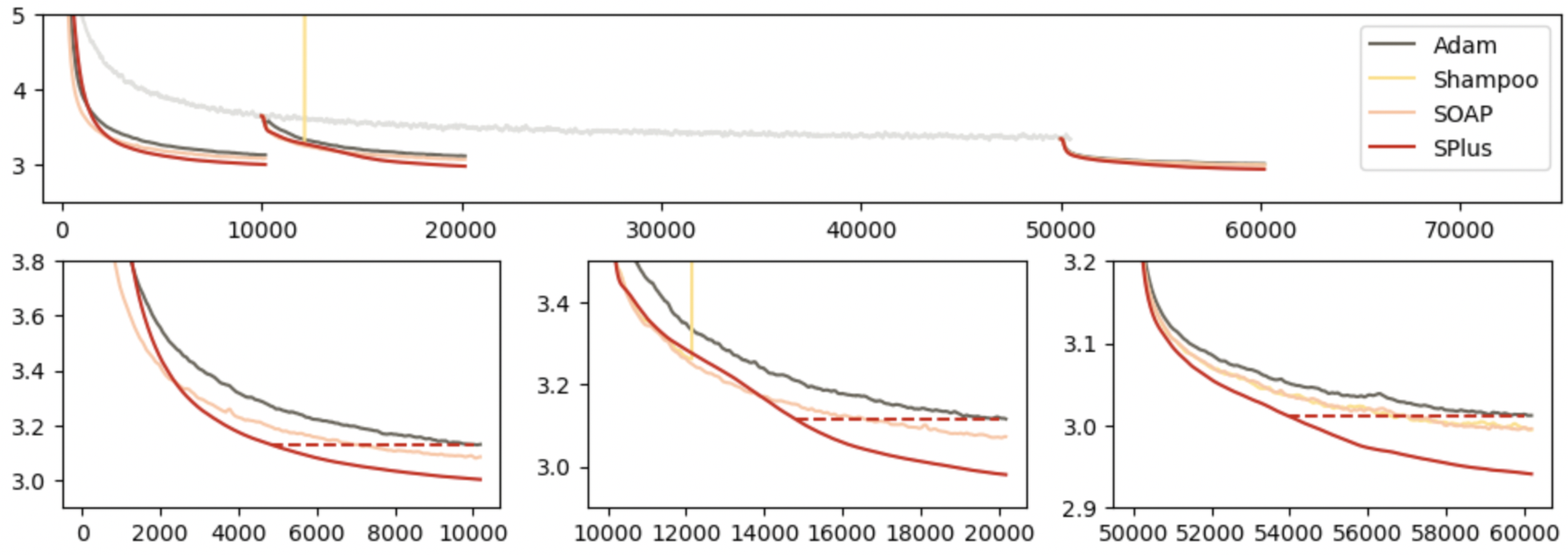}
    \caption{\textbf{Optimizers are evaluated over 10k gradient steps, starting from three distinct checkpoints per objective.} We design this setting to test robustness across objectives and across stages of training. As shown above for the LLM case, SPlus consistently reaches the same validation performance as Adam within a smaller fraction of gradient steps \maincolor{(dotted line)}.}
    \label{fig:main}
    \vspace{10.0pt}

    \begin{tabular}{lrrrrr}
    \toprule
    Method & Steps-To-Adam$^1$ & STA (LLM) & STA (ViT) & STA (DiT) & Time-to-Adam$^1$\\
    \midrule
    Naive SGD & > 10.0 & > 10.0 & > 10.0 & > 10.0 & > 10.0 \\
    Adam & 1.0 & 1.0 & 1.0 & 1.0 & 1.0 \\
    Sch.Free Adam & 0.679 & 0.674 & 0.698 & 0.664 & \textbf{0.654} \\
    Sophia & > 1.0 & > 1.0 & n/a & n/a & > 1.0 \\
    Shampoo & $0.699^{\;2}$ & 0.699 & Diverge & Diverge & $2.426^{\; 2}$ \\
    SOAP & 0.575 & 0.683 & 0.567 & 0.477 & 0.807 \\
    PSGD & 0.652 & 0.705 & 0.615 & 0.636 & 0.940 \\
    Muon & 0.832 & >1.0 & 0.920 & 0.877 & 0.934 \\
    \textbf{SPlus (ours)} & \textbf{0.439} & \textbf{0.419} & \textbf{0.504} & \textbf{0.396} & \textbf{0.617} \\
    \bottomrule
    \end{tabular}
    \vspace{0.5pt}
    \newline
    $^1$ Average values over LLM, ViT, and DiT. $\qquad ^2$ Only considering non-divergent settings.
    \vspace{0.5pt}
    \caption{\textbf{SPlus outperforms prior methods in both gradient steps and wallclock time, and matches Adam performance within \maincolor{$\mathbf{44-58\%}$} of the gradient steps.} Learning rates are swept independently for each method. We examine the training of a 160M-parameter Transformer with a batch size of 1024, and a sequence/patch length of 256. Results are averaged starting from three base checkpoints, and the full results are in \cref{table:fullresults1} of the Appendix.}
    \label{table:comparison}
\end{figure}

To thoroughly compare between optimizer types, we consider the performance across different stages of training. Concretely, we construct base checkpoints by with Adam, and saving checkpoints at fixed intervals (initialization, ten thousand, and fifty thousand steps). We then evaluate each optimizer on the three checkpoints, training for ten thousand additional gradient steps. Learning rate is swept independently for each optimizer type, along a resolution of $10^{1/3}$, e.g. $(0.0001, 0.000215, 0.000464, 0.001, ...)$. Final performance is reported as validation loss after this procedure, measured on a fixed held-out validation set. The same random seed and data order are used in each run.

As loss scales vary per objective, we focus on \textit{steps-to-Adam} as the main metric. We record the fraction of gradient steps and fraction of wallclock time required to match the performance of Adam on the task, measured via validation loss. Wall-clock results are machine-specific and should be seen as a rough estimate; we run all experiments on the same set of 32 TPUv3 pods, a typical run takes half a day. Results are reported utilizing the best-performing learning rate for each optimizer. 

We measure performance against the following optimizers which broadly span the literature:
\begin{itemize}
    \item \textbf{Naive SGD}, which does not modify the gradient except for scaling by a global learning rate.
    \item \textbf{Adam} \citep{kingma2014adam}, the main baseline, which keeps track of an elementwise uncentered variance, then scales the gradient elementwise.
    \item \textbf{Schedule-Free Adam} \citep{defazio2024road}, which replaces traditional momentum with a set of live and slow parameters. Gradients are evaluated at a linear interpolation of the two.
    \item \textbf{Sophia} \citep{liu2023sophia}, which computes an elementwise estimate of the Hessian, then scales the gradient followed by a clipping step. Sophia requires auxiliary backwards passes for the Hessian calculation, and caches the results for 10 steps.
    \item \textbf{Shampoo} \citep{gupta2018shampoo}, as described in the background section. We do not use learning-rate grafting. Matrix inversion is performed every 10 steps (and otherwise diverges). We additionally consider Shampoo with a 1/2 power, following \citep{morwani2024new, shi2023distributed}.
    \item \textbf{SOAP} \citep{vyas2024soap}, a variant on Shampoo which also tracks an elementwise uncentered variance, akin to running Adam in the eigenbasis of Shampoo. Matrix inversion is performed every 100 steps.
    \item \textbf{PSGD} \citep{li2017preconditioned}, which keeps track of an inverse whitening matrix calculated via iterated gradient descent rather than explicit matrix inversion. We use the Kron version which is known to perform the best.
    \item \textbf{Muon} \citep{jordanmuon}, which performs an orthogonalization procedure on each gradient via Newton-Schulz iteration at every update, without historical information. As Muon utilizes a higher inherent learning rate, we use a weight decay of $0.001$.
\end{itemize}

As displayed in \cref{table:comparison}, SPlus is able to outperform prior methods across the board in both gradient steps and wallclock time. We find that a well-tuned Adam is a hard baseline to beat. For example, we were unable to match the performance of Adam with Sophia (a similar finding was reported in \citep{zhao2024deconstructing}) or with Muon for LLM training. Shampoo was especially unstable (as discussed in earlier sections), and we find that when divergence does not occur, Shampoo training curves roughly match those of SOAP, as also reported in \citep{vyas2024soap}. In terms of wall-clock performance, a strong contender is Schedule-Free Adam, which does not perform any matrix-based computation and only utilizes elementwise operators. For the main table, we did not heavily tune the matrix-inversion frequency for SPlus or SOAP, and assign it a default value of $100$ -- \cref{table:frequency} shows the effect of ablating this frequency.

\subsection{What implementation details matter for SPlus?}

\begin{table*}
    \centering
    \begin{tabular}{lrrrrrrr}
    \toprule
    For frequency $\rightarrow$ & 1 & 5 & 10 & 25 & 100 & 250 & 500\\
    \midrule
    Steps-to-Adam, SPlus & 0.78	& \textbf{0.71} &	0.75 &	0.76 &	0.77 &	0.77 &	0.81 \\
    Steps-to-Adam, Shampoo & 0.86 &	0.80 &	0.80 &	0.79 & Div. &	Div. &	Div. \\
    \midrule
    Time per Update, SPlus & 7.39 &	1.79 &	1.09	 & 0.70 &	0.48 &	0.44 &	0.43 \\
    Time per Update, Shampoo & 7.41 &	1.78 &	1.09	 & 0.67 &	0.42	& 0.41	& 0.40 \\
    Time per Update, Adam & 0.34 \\
    \midrule
    Time-to-Adam, SPlus & 16.21 &	3.34	& 2.05	& 1.20	& 0.74 &	\textbf{0.65} &	0.67 \\
    Time-to-Adam, Shampoo & 18.11	& 3.79	& 2.18	& 1.21	& Div.	& Div.	& Div \\
    \bottomrule
    \end{tabular}
    \vspace{0.5pt}
    \caption{\textbf{Eigendecomposition frequency trades off between performance and wallclock time.} Each trial is run for 1000 gradient steps on the LLM-Init objective. Note that specific wallclock times are machine-dependent (we train on a v3-32 TPU) and may vary between systems.}
    \label{table:frequency}
\end{table*}

\begin{table*}
    \begin{tabular}{lrrrr}
    \toprule
    Method & LLM-Init & LLM-10K & LLM-50K & Memory Usage \\
    \midrule
    Adam & 1.0 & 1.0 & 1.0 & $3mn$ \\
    S.Free Adam & > 1.0 & 0.532 & 0.49 & $3mn$ \\
    Adam + Iterate Averaging & 0.677 & 0.52 & 0.81 & $4mn$ \\
    Shampoo & Div. & Div. & 0.699 & $2mn + 2(m^2+n^2)$ \\
    SPlus (No Averaging) & 0.729 & 0.662 & 0.685 & $2mn + 2(m^2+n^2)$ \\
    SPlus & \textbf{0.487} & \textbf{0.422} & 0.348 & $3mn + 2(m^2+n^2)$ \\
    SOAP & 0.712 & 0.66 & 0.677 & $3mn + 2(m^2+n^2)$ \\
    SOAP + Iterate Averaging & 0.513 & \textbf{0.416} & \textbf{0.317} & $4mn + 2(m^2+n^2)$ \\
    \bottomrule
    \end{tabular}
    \caption{Ablating the usage of iterate averaging vs. the underlying gradient transformation.}
    \label{table:iterate}
\end{table*}

We only apply the SPlus update on two-dimensional dense layers, which in the case of a Transformer, composes a majority of the backbone (the exception being LayerNorm scale parameters). We also do not apply SPlus to the domain-specific input and output layers -- e.g. the token embedding, the classification head, and the convolutional patch layers. For these nonstandard parameters, we simply set the update as the sign of the momentum values. Additionally, for nonstandard parameters where the shape-dependent scaling term of \cref{eq:scaling} is undefined, we use a fixed constant scaling ($0.001$ in our experiments). We found that this constant is not sensitive to even $10$x or $0.1$x pertubations, and does not need to be tuned.

We perform the above experiments over a pod of 32 TPUv3 machines, and parameters are distributed in a fully-sharded data parallel \citep{zhao2023pytorch} setup. The SPlus update is distributed among devices. Specifically, the per-step computations are calculated independently on each device as usual. However, for the eigendecomposition which occurs every N steps, we instead broadcast the $L$ and $R$ matrices evenly among devices. In parallel, each device then performs the eigendecomposition for its assigned matrices, then re-broadcasts the results. In this way, the most expensive step of the SPlus update can be reduced by a factor of $(1/\text{num devices})$.

To ablate the specific changes proposed in SPlus, we show in \cref{table:iterate} an ablation on the interaction between iterate-averaging and instant-sign normalization, focusing specifically on the LLM setting. Iterate averaging is consistently helpful regardless of the underlying gradient transformation. This trend may be a result of our usage of a constant learning rate; as discussed in \citep{defazio2024road, izmailov2018averaging}, iterate averaging can be seen as a stand-in for learning rate decay. The downside is the memory requirement for an additional set of parameters. We note that SOAP + Iterate Averaging is competetive and can outperform SPlus. One way to contrast these methods is that the instant-sign normalization used in SPlus is a memory-efficient approximation of the internal Adam optimizer used in SOAP, which otherwise requires an additional set of parameters in memory.

\section{Discussion and Conclusion}
In this work, we present SPlus, a stable whitening optimizer for neural network training. Through a fundamentally experimental approach, we introduce three key changes to improve scalability. First, direct multiplication by the square-root inverse is replaced by instant-sign normalization, which dramatically improves stability. Second, updates are correctly scaled for learning rate to transfer among network widths. Third, iterate averaging is applied to the live parameters, which reduces parameter noise and enables using a larger learning rate.

Empirically, we show that SPlus can achieve the same validation performance as Adam with $\sim 44$-$58\%$ of the gradient steps, and in $\sim 62$-$83\%$ of the wall-clock time. Over a range of training objectives and  checkpoints, SPlus achieves superior performance in comparison to previous optimizers.

\textbf{Code.} We provide full open-source code to replicate these experiments at \href{https://github.com/kvfrans/splus}{github.com/kvfrans/splus}.

\textbf{Limitations.} In comparison to Adam, SPlus requires storing $3nm + 2(n^2+m^2)$ parameters per dense layer--three instances for the live parameters, slow parameters, and momentum, along with two Kronecker factors for the gradient covariances plus the cached eigenvectors. For a square matrix, this is roughly $60\%$ more memory than Adam. Furthermore, SPlus requires nontrivial wall-clock time for matrix eigendecomposition. We note that SPlus (without iterate-averaging) uses the same amount of compute as Shampoo, and less memory than SOAP. The training settings in this work only consider Transformer architectures. While this is an intentional choice, as neural networks today have largely converged on the Transformer backbone, it remains unanswered how performance would vary on non-Transformer architectures. We utilize a constant learning rate as is common in diffusion model training \citep{peebles2023scalable}, but is less standard in language model training.

\textbf{Future directions.} We believe that SPlus, as well as our scientific setup as a whole, opens up directions in scalable neural network optimizers. We are curious on the results of applying SPlus to large-scale training at the billion-parameter scale. By providing a thorough evaluation setup, we hope to lower the bar of experimenting with new strategies, including extensions to SPlus. For example, alternate factorizations or low-rank approximations could reduce the computational cost of whitening, and a strategy may exist in-between whitening and Hessian-based conditioning. Such ideas should be evaluated in a reproducible way, following the methodology developed here. On a practical level, we hope the efficiency and simplicity of SPlus allows the community to easily plug-and-play into their desired training objectives, and train neural networks faster as a whole.

\section{Acknowledgments}

This work was supported in part by an NSF Fellowship for KF, under grant No. DGE 2146752. Any opinions, findings, and conclusions or recommendations expressed in this material are those of the author(s) and do not necessarily reflect the views of the NSF. PA holds concurrent appointments as a Professor at UC Berkeley and as an Amazon Scholar. This paper describes work performed at UC Berkeley and is not associated with Amazon. We thank Google TPU Research Cloud (TRC) for granting us access to TPUs for research.

\bibliography{iclr2023_conference}

\begin{thebibliography}{59}
\providecommand{\natexlab}[1]{#1}
\providecommand{\url}[1]{\texttt{#1}}
\expandafter\ifx\csname urlstyle\endcsname\relax
  \providecommand{\doi}[1]{doi: #1}\else
  \providecommand{\doi}{doi: \begingroup \urlstyle{rm}\Url}\fi

\bibitem[Achiam et~al.(2023)Achiam, Adler, Agarwal, Ahmad, Akkaya, Aleman, Almeida, Altenschmidt, Altman, Anadkat, et~al.]{achiam2023gpt}
Josh Achiam, Steven Adler, Sandhini Agarwal, Lama Ahmad, Ilge Akkaya, Florencia~Leoni Aleman, Diogo Almeida, Janko Altenschmidt, Sam Altman, Shyamal Anadkat, et~al.
\newblock Gpt-4 technical report.
\newblock \emph{arXiv preprint arXiv:2303.08774}, 2023.

\bibitem[Agarwal et~al.(2020)Agarwal, Anil, Hazan, Koren, and Zhang]{agarwal2020disentangling}
Naman Agarwal, Rohan Anil, Elad Hazan, Tomer Koren, and Cyril Zhang.
\newblock Disentangling adaptive gradient methods from learning rates.
\newblock \emph{arXiv preprint arXiv:2002.11803}, 2020.

\bibitem[Amari(1998)]{amari1998natural}
Shun-Ichi Amari.
\newblock Natural gradient works efficiently in learning.
\newblock \emph{Neural computation}, 10\penalty0 (2):\penalty0 251--276, 1998.

\bibitem[Anil et~al.(2020)Anil, Gupta, Koren, Regan, and Singer]{anil2020scalable}
Rohan Anil, Vineet Gupta, Tomer Koren, Kevin Regan, and Yoram Singer.
\newblock Scalable second order optimization for deep learning.
\newblock \emph{arXiv preprint arXiv:2002.09018}, 2020.

\bibitem[Bernstein \& Newhouse(2024)Bernstein and Newhouse]{bernstein2024old}
Jeremy Bernstein and Laker Newhouse.
\newblock Old optimizer, new norm: An anthology.
\newblock \emph{arXiv preprint arXiv:2409.20325}, 2024.

\bibitem[Bernstein et~al.(2018)Bernstein, Wang, Azizzadenesheli, and Anandkumar]{bernstein2018signsgd}
Jeremy Bernstein, Yu-Xiang Wang, Kamyar Azizzadenesheli, and Animashree Anandkumar.
\newblock signsgd: Compressed optimisation for non-convex problems.
\newblock In \emph{International Conference on Machine Learning}, pp.\  560--569. PMLR, 2018.

\bibitem[Bottou et~al.(2018)Bottou, Curtis, and Nocedal]{bottou2018optimization}
L{\'e}on Bottou, Frank~E Curtis, and Jorge Nocedal.
\newblock Optimization methods for large-scale machine learning.
\newblock \emph{SIAM review}, 60\penalty0 (2):\penalty0 223--311, 2018.

\bibitem[Brown et~al.(2020)Brown, Mann, Ryder, Subbiah, Kaplan, Dhariwal, Neelakantan, Shyam, Sastry, Askell, et~al.]{brown2020language}
Tom Brown, Benjamin Mann, Nick Ryder, Melanie Subbiah, Jared~D Kaplan, Prafulla Dhariwal, Arvind Neelakantan, Pranav Shyam, Girish Sastry, Amanda Askell, et~al.
\newblock Language models are few-shot learners.
\newblock \emph{Advances in neural information processing systems}, 33:\penalty0 1877--1901, 2020.

\bibitem[Defazio et~al.(2024)Defazio, Yang, Khaled, Mishchenko, Mehta, and Cutkosky]{defazio2024road}
Aaron Defazio, Xingyu Yang, Ahmed Khaled, Konstantin Mishchenko, Harsh Mehta, and Ashok Cutkosky.
\newblock The road less scheduled.
\newblock \emph{Advances in Neural Information Processing Systems}, 37:\penalty0 9974--10007, 2024.

\bibitem[Dosovitskiy et~al.(2020)Dosovitskiy, Beyer, Kolesnikov, Weissenborn, Zhai, Unterthiner, Dehghani, Minderer, Heigold, Gelly, et~al.]{dosovitskiy2020image}
Alexey Dosovitskiy, Lucas Beyer, Alexander Kolesnikov, Dirk Weissenborn, Xiaohua Zhai, Thomas Unterthiner, Mostafa Dehghani, Matthias Minderer, Georg Heigold, Sylvain Gelly, et~al.
\newblock An image is worth 16x16 words: Transformers for image recognition at scale.
\newblock \emph{arXiv preprint arXiv:2010.11929}, 2020.

\bibitem[Duchi et~al.(2011)Duchi, Hazan, and Singer]{duchi2011adaptive}
John Duchi, Elad Hazan, and Yoram Singer.
\newblock Adaptive subgradient methods for online learning and stochastic optimization.
\newblock \emph{Journal of machine learning research}, 12\penalty0 (7), 2011.

\bibitem[Fujimoto et~al.(2018)Fujimoto, Hoof, and Meger]{fujimoto2018addressing}
Scott Fujimoto, Herke Hoof, and David Meger.
\newblock Addressing function approximation error in actor-critic methods.
\newblock In \emph{International conference on machine learning}, pp.\  1587--1596. PMLR, 2018.

\bibitem[Gokaslan et~al.(2019)Gokaslan, Cohen, Pavlick, and Tellex]{Gokaslan2019OpenWeb}
Aaron Gokaslan, Vanya Cohen, Ellie Pavlick, and Stefanie Tellex.
\newblock Openwebtext corpus.
\newblock \url{http://Skylion007.github.io/OpenWebTextCorpus}, 2019.

\bibitem[Grill et~al.(2020)Grill, Strub, Altch{\'e}, Tallec, Richemond, Buchatskaya, Doersch, Avila~Pires, Guo, Gheshlaghi~Azar, et~al.]{grill2020bootstrap}
Jean-Bastien Grill, Florian Strub, Florent Altch{\'e}, Corentin Tallec, Pierre Richemond, Elena Buchatskaya, Carl Doersch, Bernardo Avila~Pires, Zhaohan Guo, Mohammad Gheshlaghi~Azar, et~al.
\newblock Bootstrap your own latent-a new approach to self-supervised learning.
\newblock \emph{Advances in neural information processing systems}, 33:\penalty0 21271--21284, 2020.

\bibitem[Gupta et~al.(2018)Gupta, Koren, and Singer]{gupta2018shampoo}
Vineet Gupta, Tomer Koren, and Yoram Singer.
\newblock Shampoo: Preconditioned stochastic tensor optimization.
\newblock In \emph{International Conference on Machine Learning}, pp.\  1842--1850. PMLR, 2018.

\bibitem[Ho et~al.(2020)Ho, Jain, and Abbeel]{ho2020denoising}
Jonathan Ho, Ajay Jain, and Pieter Abbeel.
\newblock Denoising diffusion probabilistic models.
\newblock \emph{Advances in neural information processing systems}, 33:\penalty0 6840--6851, 2020.

\bibitem[Izmailov et~al.(2018)Izmailov, Podoprikhin, Garipov, Vetrov, and Wilson]{izmailov2018averaging}
Pavel Izmailov, Dmitrii Podoprikhin, Timur Garipov, Dmitry Vetrov, and Andrew~Gordon Wilson.
\newblock Averaging weights leads to wider optima and better generalization.
\newblock \emph{arXiv preprint arXiv:1803.05407}, 2018.

\bibitem[Jordan et~al.()Jordan, Jin, Boza, Jiacheng, Cecista, Newhouse, and Bernstein]{jordanmuon}
K~Jordan, Y~Jin, V~Boza, Y~Jiacheng, F~Cecista, L~Newhouse, and J~Bernstein.
\newblock Muon: An optimizer for hidden layers in neural networks, 2024b.
\newblock \emph{URL https://kellerjordan. github. io/posts/muon}.

\bibitem[Karras et~al.(2024)Karras, Aittala, Lehtinen, Hellsten, Aila, and Laine]{karras2024analyzing}
Tero Karras, Miika Aittala, Jaakko Lehtinen, Janne Hellsten, Timo Aila, and Samuli Laine.
\newblock Analyzing and improving the training dynamics of diffusion models.
\newblock In \emph{Proceedings of the IEEE/CVF Conference on Computer Vision and Pattern Recognition}, pp.\  24174--24184, 2024.

\bibitem[Kingma \& Ba(2014)Kingma and Ba]{kingma2014adam}
Diederik~P Kingma and Jimmy Ba.
\newblock Adam: A method for stochastic optimization.
\newblock \emph{arXiv preprint arXiv:1412.6980}, 2014.

\bibitem[Korbit et~al.(2024)Korbit, Adeoye, Bemporad, and Zanon]{korbit2024exact}
Mikalai Korbit, Adeyemi~D Adeoye, Alberto Bemporad, and Mario Zanon.
\newblock Exact gauss-newton optimization for training deep neural networks.
\newblock \emph{arXiv preprint arXiv:2405.14402}, 2024.

\bibitem[Kunstner et~al.(2019)Kunstner, Hennig, and Balles]{kunstner2019limitations}
Frederik Kunstner, Philipp Hennig, and Lukas Balles.
\newblock Limitations of the empirical fisher approximation for natural gradient descent.
\newblock \emph{Advances in neural information processing systems}, 32, 2019.

\bibitem[Large et~al.(2024)Large, Liu, Huh, Bahng, Isola, and Bernstein]{large2024scalable}
Tim Large, Yang Liu, Minyoung Huh, Hyojin Bahng, Phillip Isola, and Jeremy Bernstein.
\newblock Scalable optimization in the modular norm.
\newblock \emph{arXiv preprint arXiv:2405.14813}, 2024.

\bibitem[LeCun et~al.(2002)LeCun, Bottou, Orr, and M{\"u}ller]{lecun2002efficient}
Yann LeCun, L{\'e}on Bottou, Genevieve~B Orr, and Klaus-Robert M{\"u}ller.
\newblock Efficient backprop.
\newblock In \emph{Neural networks: Tricks of the trade}, pp.\  9--50. Springer, 2002.

\bibitem[Li(2017)]{li2017preconditioned}
Xi-Lin Li.
\newblock Preconditioned stochastic gradient descent.
\newblock \emph{IEEE transactions on neural networks and learning systems}, 29\penalty0 (5):\penalty0 1454--1466, 2017.

\bibitem[Lipman et~al.(2022)Lipman, Chen, Ben-Hamu, Nickel, and Le]{lipman2022flow}
Yaron Lipman, Ricky~TQ Chen, Heli Ben-Hamu, Maximilian Nickel, and Matt Le.
\newblock Flow matching for generative modeling.
\newblock \emph{arXiv preprint arXiv:2210.02747}, 2022.

\bibitem[Liu et~al.(2023)Liu, Li, Hall, Liang, and Ma]{liu2023sophia}
Hong Liu, Zhiyuan Li, David Hall, Percy Liang, and Tengyu Ma.
\newblock Sophia: A scalable stochastic second-order optimizer for language model pre-training.
\newblock \emph{arXiv preprint arXiv:2305.14342}, 2023.

\bibitem[Loshchilov \& Hutter(2017)Loshchilov and Hutter]{loshchilov2017decoupled}
Ilya Loshchilov and Frank Hutter.
\newblock Decoupled weight decay regularization.
\newblock \emph{arXiv preprint arXiv:1711.05101}, 2017.

\bibitem[Ma et~al.(2024{\natexlab{a}})Ma, Gong, Scetbon, and Meeds]{ma2024swan}
Chao Ma, Wenbo Gong, Meyer Scetbon, and Edward Meeds.
\newblock Swan: Preprocessing sgd enables adam-level performance on llm training with significant memory reduction.
\newblock \emph{arXiv preprint arXiv:2412.13148}, 2024{\natexlab{a}}.

\bibitem[Ma et~al.(2024{\natexlab{b}})Ma, Wang, Jia, Chen, Liu, Li, Chen, and Qiao]{ma2024latte}
Xin Ma, Yaohui Wang, Gengyun Jia, Xinyuan Chen, Ziwei Liu, Yuan-Fang Li, Cunjian Chen, and Yu~Qiao.
\newblock Latte: Latent diffusion transformer for video generation.
\newblock \emph{arXiv preprint arXiv:2401.03048}, 2024{\natexlab{b}}.

\bibitem[Martens(2020)]{martens2020new}
James Martens.
\newblock New insights and perspectives on the natural gradient method.
\newblock \emph{Journal of Machine Learning Research}, 21\penalty0 (146):\penalty0 1--76, 2020.

\bibitem[Martens \& Grosse(2015)Martens and Grosse]{martens2015optimizing}
James Martens and Roger Grosse.
\newblock Optimizing neural networks with kronecker-factored approximate curvature.
\newblock In \emph{International conference on machine learning}, pp.\  2408--2417. PMLR, 2015.

\bibitem[Martens et~al.(2010)]{martens2010deep}
James Martens et~al.
\newblock Deep learning via hessian-free optimization.
\newblock In \emph{Icml}, volume~27, pp.\  735--742, 2010.

\bibitem[McCandlish et~al.(2018)McCandlish, Kaplan, Amodei, and Team]{mccandlish2018empirical}
Sam McCandlish, Jared Kaplan, Dario Amodei, and OpenAI~Dota Team.
\newblock An empirical model of large-batch training.
\newblock \emph{arXiv preprint arXiv:1812.06162}, 2018.

\bibitem[Morales-Brotons et~al.(2024)Morales-Brotons, Vogels, and Hendrikx]{morales2024exponential}
Daniel Morales-Brotons, Thijs Vogels, and Hadrien Hendrikx.
\newblock Exponential moving average of weights in deep learning: Dynamics and benefits.
\newblock \emph{arXiv preprint arXiv:2411.18704}, 2024.

\bibitem[Morwani et~al.(2024)Morwani, Shapira, Vyas, Malach, Kakade, and Janson]{morwani2024new}
Depen Morwani, Itai Shapira, Nikhil Vyas, Eran Malach, Sham Kakade, and Lucas Janson.
\newblock A new perspective on shampoo's preconditioner.
\newblock \emph{arXiv preprint arXiv:2406.17748}, 2024.

\bibitem[Peebles \& Xie(2023)Peebles and Xie]{peebles2023scalable}
William Peebles and Saining Xie.
\newblock Scalable diffusion models with transformers.
\newblock In \emph{Proceedings of the IEEE/CVF international conference on computer vision}, pp.\  4195--4205, 2023.

\bibitem[Podell et~al.(2023)Podell, English, Lacey, Blattmann, Dockhorn, M{\"u}ller, Penna, and Rombach]{podell2023sdxl}
Dustin Podell, Zion English, Kyle Lacey, Andreas Blattmann, Tim Dockhorn, Jonas M{\"u}ller, Joe Penna, and Robin Rombach.
\newblock Sdxl: Improving latent diffusion models for high-resolution image synthesis.
\newblock \emph{arXiv preprint arXiv:2307.01952}, 2023.

\bibitem[Polyak(1990)]{polyak1990new}
Boris~T Polyak.
\newblock New stochastic approximation type procedures.
\newblock \emph{Automat. i Telemekh}, 7\penalty0 (98-107):\penalty0 2, 1990.

\bibitem[Polyak \& Juditsky(1992)Polyak and Juditsky]{polyak1992acceleration}
Boris~T Polyak and Anatoli~B Juditsky.
\newblock Acceleration of stochastic approximation by averaging.
\newblock \emph{SIAM journal on control and optimization}, 30\penalty0 (4):\penalty0 838--855, 1992.

\bibitem[Pooladzandi \& Li(2024)Pooladzandi and Li]{pooladzandi2024curvature}
Omead Pooladzandi and Xi-Lin Li.
\newblock Curvature-informed sgd via general purpose lie-group preconditioners.
\newblock \emph{arXiv preprint arXiv:2402.04553}, 2024.

\bibitem[Radford et~al.(2019)Radford, Wu, Child, Luan, Amodei, Sutskever, et~al.]{radford2019language}
Alec Radford, Jeffrey Wu, Rewon Child, David Luan, Dario Amodei, Ilya Sutskever, et~al.
\newblock Language models are unsupervised multitask learners.
\newblock \emph{OpenAI blog}, 1\penalty0 (8):\penalty0 9, 2019.

\bibitem[Rombach et~al.(2022)Rombach, Blattmann, Lorenz, Esser, and Ommer]{rombach2022high}
Robin Rombach, Andreas Blattmann, Dominik Lorenz, Patrick Esser, and Bj{\"o}rn Ommer.
\newblock High-resolution image synthesis with latent diffusion models.
\newblock In \emph{Proceedings of the IEEE/CVF conference on computer vision and pattern recognition}, pp.\  10684--10695, 2022.

\bibitem[Ruppert(1988)]{ruppert1988efficient}
David Ruppert.
\newblock Efficient estimations from a slowly convergent robbins-monro process.
\newblock Technical report, Cornell University Operations Research and Industrial Engineering, 1988.

\bibitem[Schraudolph(2002)]{schraudolph2002fast}
Nicol~N Schraudolph.
\newblock Fast curvature matrix-vector products for second-order gradient descent.
\newblock \emph{Neural computation}, 14\penalty0 (7):\penalty0 1723--1738, 2002.

\bibitem[Shazeer \& Stern(2018)Shazeer and Stern]{shazeer2018adafactor}
Noam Shazeer and Mitchell Stern.
\newblock Adafactor: Adaptive learning rates with sublinear memory cost.
\newblock In \emph{International Conference on Machine Learning}, pp.\  4596--4604. PMLR, 2018.

\bibitem[Shi et~al.(2023)Shi, Lee, Iwasaki, Gallego-Posada, Li, Rangadurai, Mudigere, and Rabbat]{shi2023distributed}
Hao-Jun~Michael Shi, Tsung-Hsien Lee, Shintaro Iwasaki, Jose Gallego-Posada, Zhijing Li, Kaushik Rangadurai, Dheevatsa Mudigere, and Michael Rabbat.
\newblock A distributed data-parallel pytorch implementation of the distributed shampoo optimizer for training neural networks at-scale.
\newblock \emph{arXiv preprint arXiv:2309.06497}, 2023.

\bibitem[Smith et~al.(2017)Smith, Kindermans, Ying, and Le]{smith2017don}
Samuel~L Smith, Pieter-Jan Kindermans, Chris Ying, and Quoc~V Le.
\newblock Don't decay the learning rate, increase the batch size.
\newblock \emph{arXiv preprint arXiv:1711.00489}, 2017.

\bibitem[Sohl-Dickstein(2012)]{sohl2012natural}
Jascha Sohl-Dickstein.
\newblock The natural gradient by analogy to signal whitening, and recipes and tricks for its use.
\newblock \emph{arXiv preprint arXiv:1205.1828}, 2012.

\bibitem[Team et~al.(2024)Team, Ghosh, Walke, Pertsch, Black, Mees, Dasari, Hejna, Kreiman, Xu, et~al.]{team2024octo}
Octo~Model Team, Dibya Ghosh, Homer Walke, Karl Pertsch, Kevin Black, Oier Mees, Sudeep Dasari, Joey Hejna, Tobias Kreiman, Charles Xu, et~al.
\newblock Octo: An open-source generalist robot policy.
\newblock \emph{arXiv preprint arXiv:2405.12213}, 2024.

\bibitem[Van~Hasselt et~al.(2016)Van~Hasselt, Guez, and Silver]{van2016deep}
Hado Van~Hasselt, Arthur Guez, and David Silver.
\newblock Deep reinforcement learning with double q-learning.
\newblock In \emph{Proceedings of the AAAI conference on artificial intelligence}, volume~30, 2016.

\bibitem[Vaswani et~al.(2017)Vaswani, Shazeer, Parmar, Uszkoreit, Jones, Gomez, Kaiser, and Polosukhin]{vaswani2017attention}
Ashish Vaswani, Noam Shazeer, Niki Parmar, Jakob Uszkoreit, Llion Jones, Aidan~N Gomez, {\L}ukasz Kaiser, and Illia Polosukhin.
\newblock Attention is all you need.
\newblock \emph{Advances in neural information processing systems}, 30, 2017.

\bibitem[Vyas et~al.(2024)Vyas, Morwani, Zhao, Kwun, Shapira, Brandfonbrener, Janson, and Kakade]{vyas2024soap}
Nikhil Vyas, Depen Morwani, Rosie Zhao, Mujin Kwun, Itai Shapira, David Brandfonbrener, Lucas Janson, and Sham Kakade.
\newblock Soap: Improving and stabilizing shampoo using adam.
\newblock \emph{arXiv preprint arXiv:2409.11321}, 2024.

\bibitem[Yang et~al.(2022)Yang, Hu, Babuschkin, Sidor, Liu, Farhi, Ryder, Pachocki, Chen, and Gao]{yang2022tensor}
Greg Yang, Edward~J Hu, Igor Babuschkin, Szymon Sidor, Xiaodong Liu, David Farhi, Nick Ryder, Jakub Pachocki, Weizhu Chen, and Jianfeng Gao.
\newblock Tensor programs v: Tuning large neural networks via zero-shot hyperparameter transfer.
\newblock \emph{arXiv preprint arXiv:2203.03466}, 2022.

\bibitem[Yang et~al.(2023)Yang, Simon, and Bernstein]{yang2023spectral}
Greg Yang, James~B Simon, and Jeremy Bernstein.
\newblock A spectral condition for feature learning.
\newblock \emph{arXiv preprint arXiv:2310.17813}, 2023.

\bibitem[Yang \& Laaksonen(2008)Yang and Laaksonen]{yang2008principal}
Zhirong Yang and Jorma Laaksonen.
\newblock Principal whitened gradient for information geometry.
\newblock \emph{Neural Networks}, 21\penalty0 (2-3):\penalty0 232--240, 2008.

\bibitem[Yaz et~al.(2018)Yaz, Foo, Winkler, Yap, Piliouras, Chandrasekhar, et~al.]{yaz2018unusual}
Yasin Yaz, Chuan-Sheng Foo, Stefan Winkler, Kim-Hui Yap, Georgios Piliouras, Vijay Chandrasekhar, et~al.
\newblock The unusual effectiveness of averaging in gan training.
\newblock In \emph{International Conference on Learning Representations}, 2018.

\bibitem[Zhao et~al.(2024)Zhao, Morwani, Brandfonbrener, Vyas, and Kakade]{zhao2024deconstructing}
Rosie Zhao, Depen Morwani, David Brandfonbrener, Nikhil Vyas, and Sham Kakade.
\newblock Deconstructing what makes a good optimizer for language models.
\newblock \emph{arXiv preprint arXiv:2407.07972}, 2024.

\bibitem[Zhao et~al.(2023)Zhao, Gu, Varma, Luo, Huang, Xu, Wright, Shojanazeri, Ott, Shleifer, et~al.]{zhao2023pytorch}
Yanli Zhao, Andrew Gu, Rohan Varma, Liang Luo, Chien-Chin Huang, Min Xu, Less Wright, Hamid Shojanazeri, Myle Ott, Sam Shleifer, et~al.
\newblock Pytorch fsdp: experiences on scaling fully sharded data parallel.
\newblock \emph{arXiv preprint arXiv:2304.11277}, 2023.

\end{thebibliography}
\bibliographystyle{iclr2023_conference}

\newpage
\appendix

\section{Appendix / supplemental material}

\subsection{Pseudocode of SPlus}

We provide here a snippet of the core components of SPlus, implemented in JAX. For a full implementation, check out the repo at \href{https://github.com/kvfrans/splus}{github.com/kvfrans/splus}.

\begin{minted}{python}
class SPlusState(NamedTuple):
  ema: chex.Array
  momentum: chex.Array
  sides: chex.Array
  q_sides: chex.Array
  step: int
  ema_rate: float

def splus_get_eval_params(state):
  ema_hat = jax.tree_map(lambda e: e / (1 - state.ema_rate ** state.step), state.ema)
  return ema_hat

def splus(
      learning_rate: base.ScalarOrSchedule,
      b1: float = 0.9,
      b2: float = 0.999,
      ema_rate: float = 0.999,
      eps: float = 1e-30,
      inverse_every: int = 100,
      nonstandard_constant: float = 0.001,
      weight_decay: float = 1e-2,
      mask: Optional[Union[Any, Callable[[base.Params], Any]]] = None,
      max_dim: int = 10000,
  ):

  def init_fn(params):
    momentum = otu.tree_zeros_like(params)
    ema = otu.tree_zeros_like(params)
    def sides_decomp(p):
      if len(p.shape) == 2:
        return [jnp.zeros((d, d)) if d < max_dim else None for d in p.shape]
      return None
    sides = jax.tree_map(sides_decomp, params)
    def qs_decomp(p):
      if len(p.shape) == 2:
        return [jnp.eye(d) if d < max_dim else None for d in p.shape]
    q_sides = jax.tree_map(qs_decomp, params)
    step = 0
    return SPlusState(ema, momentum, sides, q_sides, step, ema_rate)

  def update_sides(g, s):
    if len(g.shape) == 2:
      return [
        b2 * s[0] + (1 - b2) * g @ g.T if s[0] is not None else None,
        b2 * s[1] + (1 - b2) * g.T @ g if s[1] is not None else None,
      ]
    else:
      return None

  def rot(p, q):
    if len(p.shape) == 2:
      p = q[0].T @ p if q[0] is not None else p
      p = p @ q[1] if q[1] is not None else p
    return p
  
  def unrot(p, q):
    if len(p.shape) == 2:
      p = q[0] @ p if q[0] is not None else p
      p = p @ q[1].T if q[1] is not None else p
    return p

  @jax.jit
  def get_eigvecs(s):
    if s is None:
      return None
    _, q = jnp.linalg.eigh(s + eps * jnp.eye(s.shape[0]))
    return q
  
  def update_inverse(sides):
    q_sides = jax.tree_map(get_eigvecs, sides)
    return q_sides
  
  def update_fn(grads, state, params):
    step = state.step + 1

    # Rotate to eigenbasis, take sign, unrotate.
    momentum = jax.tree_map(lambda m, g: b1 * m + (1 - b1) * g, state.momentum, grads)
    momentum_rot = jax.tree_map(rot, momentum, state.q_sides)
    updates_rot = jax.tree_map(lambda m: jnp.sign(m), momentum_rot)
    updates = jax.tree_map(unrot, updates_rot, state.q_sides)
    sides = jax.tree_map(update_sides, grads, state.sides)
    ema = jax.tree_map(lambda e, g: ema_rate * e + (1 - ema_rate) * g, state.ema, params)

    # Every `inverse_every` steps, we update the inverse eigendecomposition.
    do_inverse = (step %
    q_sides = jax.lax.cond(do_inverse, update_inverse, lambda _ : state.q_sides, sides)

    return updates, SPlusState(ema, momentum, sides, q_sides, step, state.ema_rate)
  
  def shape_scaling(updates, state, params):
    def shape_scale(path, u):
      path_str = '/'.join([p.key for p in path])
      if len(u.shape) == 2 and u.shape[0] < max_dim and u.shape[1] < max_dim:
        scale = (1 / (u.shape[0] + u.shape[1])/2)
      else:
        scale = nonstandard_constant
      return u * scale
    return jax.tree_util.tree_map_with_path(shape_scale, updates), None
  
  splus_main = base.GradientTransformation(init_fn, update_fn)
  splus_scaling = base.GradientTransformation(lambda _ : None, shape_scaling)
  return combine.chain(
    splus_main,
    transform.add_decayed_weights(weight_decay, mask),
    transform.scale_by_learning_rate(learning_rate),
    splus_scaling
  )
\end{minted}

\subsection{Full results of optimizer comparisons}

We present in \cref{table:fullresults1} and \cref{table:fullresults2} an extended table of results, comparing optimizer performance under each specific objective and starting checkpoint. As described in \cref{sec:comparison}, all experiments are conducted using a 160M parameter Transformer model. In each plot, we compare the amount of gradient steps and/or wallclock time required to match the performance of Adam. All optimizers are trained for 10k gradient steps, and the learning rate is tuned independently. "Div." indicates that training diverges under any non-trivial learning rate. "$> 1.0$" indicates that at the 10k step mark, the method is unable to outperform Adam.

\begin{table*}[h]
\scalebox{0.8}
{
    \begin{tabular}{lrrrrrrrrr}
    \toprule
    Method & LLM-Init & LLM-10K & LLM-50K & ViT-Init & ViT-10K & ViT-50K & DiT-Init & DiT-10K & DiT-50K\\
    \midrule
    Naive SGD & > 10.0 & > 10.0 & > 10.0 & > 10.0 & > 10.0 & > 10.0 & > 10.0 & > 10.0 & > 10.0 \\
    Adam & 1.0 & 1.0 & 1.0 & 1.0 & 1.0 & 1.0 & 1.0 & 1.0 & 1.0 \\
    S.Free Adam & > 1.0 & 0.532 & 0.49 & > 1.0 & 0.629 & 0.467 & > 1.0 & 0.507 & 0.487 \\
    Sophia & > 1.0 & > 1.0 & > 1.0 & n/a & n/a & n/a & n/a & n/a & n/a \\
    Shampoo & Div. & Div. & 0.699 & > 1.0 & > 1.0 & Div. & Div. & Div. & Div. \\
    SOAP & 0.712 & 0.66 & 0.677 & 0.574 & 0.57 & 0.557 & 0.486 & 0.459 & 0.488 \\
    PSGD & 0.895 & 0.628 & 0.594 & 0.535 & 0.458 & 0.852 & 0.768 & 0.412 & 0.728 \\
    Muon & > 1.0 & > 1.0 & > 1.0 & 0.978 & 0.783 & > 1.0 & 0.92 & 0.878 & 0.833 \\
    \textbf{SPlus} & \textbf{0.487} & \textbf{0.422} & \textbf{0.348} & \textbf{0.586} & \textbf{0.475} & \textbf{0.452} & \textbf{0.459} & \textbf{0.359} & \textbf{0.371} \\
    \bottomrule
    \end{tabular}
}
    \caption{Full results comparing steps-to-Adam.}
    \label{table:fullresults1}
\end{table*}

\begin{table*}[h]
\scalebox{0.8}
{
    \begin{tabular}{lrrrrrrrrr}
    \toprule
    Method & LLM-Init & LLM-10K & LLM-50K & ViT-Init & ViT-10K & ViT-50K & DiT-Init & DiT-10K & DiT-50K\\
    \midrule
    Naive SGD & > 10.0 & > 10.0 & > 10.0 & > 10.0 & > 10.0 & > 10.0 & > 10.0 & > 10.0 & > 10.0 \\
    Adam & 1.0 & 1.0 & 1.0 & 1.0 & 1.0 & 1.0 & 1.0 & 1.0 & 1.0 \\
    S.Free Adam & > 1.0 & \textbf{0.48} & \textbf{0.44} & > 1.0 & \textbf{0.593} & \textbf{0.441} & > 1.0 & \textbf{0.475} & \textbf{0.459} \\
    Sophia & > 1.0 & > 1.0 & > 1.0 & n/a & n/a & n/a & n/a & n/a & n/a \\
    Shampoo & Div. & Div. & 2.426 & > 1.0 & > 1.0 & Div. & Div. & Div. & Div. \\
    SOAP & 0.951 & 0.864 & 0.886 & 0.844 & 0.811 & 0.78 & 0.734 & 0.676 & 0.72 \\
    PSGD & 1.08 & 0.836 & 0.808 & 0.854 & 0.8 & 1.18 & 1.11 & 0.73 & 1.064 \\
    Muon & > 1.0 & > 1.0 & > 1.0 & 0.996 & 0.79 & > 1.0 & 0.915 & 0.881 & 0.827 \\
    \textbf{SPlus} & \textbf{0.651} & 0.547 & \textbf{0.447} & \textbf{0.832} & 0.674 & 0.628 & \textbf{0.707} & 0.523 & 0.545 \\
    \bottomrule
    \end{tabular}
}
    \caption{Full results comparing wallclock-to-Adam.}
    \label{table:fullresults2}
\end{table*}

\vspace*{-1ex}
\subsection{Spectral scaling vs symmetric scaling}
\vspace*{-2ex}
\label{section:symmetric}

\begin{figure}[H]
    \centering
    \includegraphics[width=\textwidth]{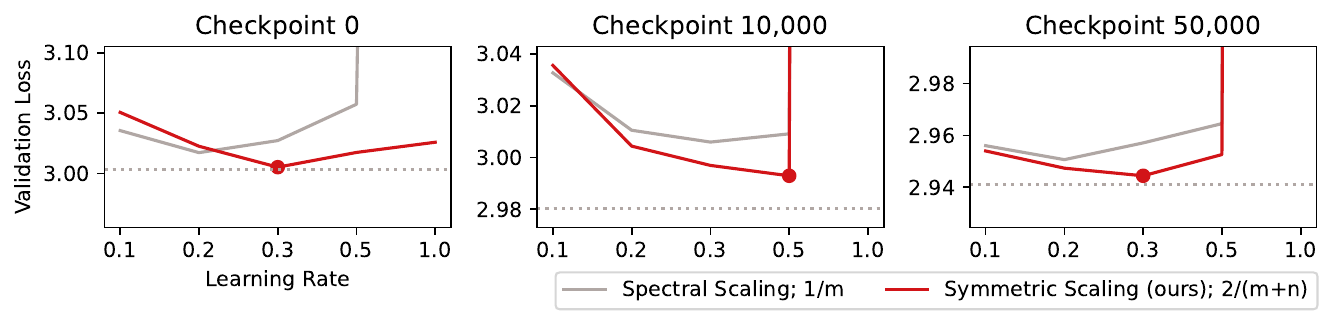}
    \caption{While both are valid strategies that enable learning rate transfer across width, we find that symmetric scaling leads to a better final performance versus spectral scaling. Dotted line shows the best-performing SPlus run \textit{without} scaling (i.e. without learning-rate transfer properties).}
    \label{fig:scaling}
    \vspace*{-2ex}
\end{figure}

In \cref{sec:scaling}, we mention a difference between the SPlus symmetric scaling factor $2/(m+n)$ versus the "spectral" scaling \citep{yang2023spectral} which argues for $1/m$. The spectral scaling is the correct factor such that regardless of \textit{any dense layer input/output ratio}, the scale of activation updates remains constant. However, we find experimentally that this property is harmful for transformer training. One hypothesis is that for non-square dense layers, gradients are effectively low-rank and/or certain eigenbases do not align with incoming activation vectors.

As shown in \cref{fig:scaling}, the symmetric scaling that we opt for maintains the original performance, outperforming spectral scaling while allowing for learning rate transfer. The rationale behind symmetric scaling is that for the core Transformer backbones -- specifically the MLP block which often consists of two layers of $(m,4m)$ and $(4m,m)$ size -- learning follows the same trajectory as if no scaling factor was used at all, i.e. a single global learning rate is applied to all parameters.

\subsection{A discussion on the Hessian, Fisher, and Empirical Fisher.}
\label{sec:preconditioning}

In this section, we provide a brief discussion on various distance metrics used in optimization methods. Recall that generalized gradient descent follows the steepest direction of improvement, where distance can be defined in terms of a metric matrix $M$:
\begin{equation}
    u = \argmin_{\Delta\theta} \; \underbrace{\; g^T\Delta\theta \;}_{\text{Improvement}} + \underbrace{(1/2)\Delta\theta^TM\Delta\theta}_{\text{Distance Penalty}} \quad = \quad M^{-1}g.
\end{equation}
where $M$ is also referred to as a \textit{preconditioner}.

\textbf{Hessian and Newton's method.} A particularly suitable choice for $M$ is the Hessian matrix, which is a matrix of second-order derivatives:
\begin{equation}
    H = E_{x\sim D} \left[ \nabla_\theta^2 L(\theta,x) \right]. 
\end{equation}
When the Hessian is used as a preconditioner, we arrive at Newton's method. Newton's method can be seen utilizing a \textit{quadratic} approximation of the loss function rather than a linear one, where the penalty for taking large steps is defined entirely by the second-order effects of that step on the loss. For this reason, Newton's method is sometimes proposed as a way to avoid tuning a learning rate, and in fact for purely quadratic loss functions, Newton's method can find the global optimum in a single iteration. 

Intuitively, one would desire that the Hessian is positive definite, such that the second-order term always results in a positive distance \textit{penalty} -- however, this is generally only true for convex loss functions. When the Hessian has negative eigenvalues, Newton's method can step in non-descending directions, or even diverge completely.

\textbf{Gauss-Newton matrix.} The Gauss-Newton matrix is an approximation to the Hessian using only first-order terms. For simplicity, let's assume the loss function is the mean-squared error of a single output. Denoting the network output as $f(\theta,x)$, we can expand the Hessian for a single $x,y$ pair:
\begin{equation}
    \nabla_\theta^2 L(\theta,x) = \underbrace{\nabla_\theta f(\theta,x) \nabla_\theta f(\theta,x)^T}_{\text{Gauss-Newton term}} + \underbrace{(f_\theta(\theta,x) - y) \nabla^2_\theta f(\theta,x)}_{\text{Dropped second-order term}}.
    \label{eq:gn}
\end{equation}
The Gauss-Newton approximation is often desired as it is strictly positive semi-definite, avoiding negative distance issues that the full Hessian has. Additionally, the Gauss-Newton term is simple to calculate as it only requires first-order gradients. The Gauss-Newton can be generalized to non-MSE losses \citep{korbit2024exact, bottou2018optimization, schraudolph2002fast} by introducing a PSD matrix $A$ between the two gradient terms:
\begin{equation}
    G = E_{x\sim D} \left[ \nabla_\theta f(\theta,x) \; A_x \; \nabla_\theta f(\theta,x)^T \right]
\end{equation}
\textbf{Fisher information matrix and natural gradient descent.} For neural networks defining probability distributions, we can use a metric that is particular to the distribution itself:
\begin{equation}
    F = E_{x\sim D, y \sim p_\theta(\cdot|x)} \left[ \nabla_\theta \log p_\theta(y|x) \nabla_\theta \log p_\theta(y|x)^T \right]
    \label{eq:fisher}
\end{equation}
which is known as the \textit{Fisher information matrix}. The Fisher does \textit{not take the loss function into account}. It is only affected by the shape of the probability distribution itself, as defined by the current neural network. When descent is performed using the Fisher as a preconditioner, it is often referred to as \textit{natural} gradient descent \citep{amari1998natural, sohl2012natural}. 

Natural gradient descent has the nice property that it is invariant to the parameterization of the network -- to a first order, optimization will follow the same trajectory regardless of a neural networks' internal structure. This property is also true for Newton's method under affine transformations of parameters.

Note the expectation in \cref{eq:fisher}, which notes that $y$ must be sampled from \textit{the current distribution.} This means that the Fisher cannot be calculated by taking the loss over samples from the dataset, and must instead use \textit{sampled} outputs.

The Fisher can also be interpreted as the Hessian of a particular loss function, namely the expectation of log-likelihood under sampled outputs:
\begin{equation}
    L_\text{Fisher} = E_{x\sim D, y \sim p_\theta(\cdot|x)} \left[  \log p_\theta(y|x) \right]
\end{equation}
where notably the second-order terms in the form of \cref{eq:gn} evaluate to zero.

\newpage
\textbf{Empirical Fisher.} An approximation often used in practice is to calculate a Fisher-like matrix, but over \textit{dataset} labels:
\begin{equation}
        F = E_{x,y\sim D} \left[ \nabla_\theta \log p_\theta(y|x) \nabla_\theta \log p_\theta(y|x)^T \right].
\end{equation}
This should not be confused with the true Fisher, as studied in \citep{mccandlish2018empirical}. In fact, the empirical Fisher is closer in nature to the Gauss-Newton matrix, and is equivalent to a generalized Gauss-Newton matrix where the inner PSD matrix is constructed as $A =\nabla^2_f \log p(y|f(\theta,x)$.

The empirical Fisher can be seen as an (uncentered) covariance of gradients. The Fisher itself is actually a \textit{centered} covariance, since the expectation of log-likelihood gradients under sampled outputs is zero. In practice, we found that centering the empirical Fisher held no practical difference.

\textbf{Whitening metric.} The \textit{whitening metric} \citep{yang2008principal} is the matrix square-root of the empirical Fisher:
\begin{equation}
    W =  E_{x,y\sim D} \left[ \nabla_\theta \log p_\theta(y|x) \nabla_\theta \log p_\theta(y|x)^T \right]^{1/2}
\end{equation}
This metric is widely used in neural network training, for example, Adam is a diagonal estimate of the whitening metric, and Shampoo is a Kronecker approximation of this term. The name \textit{whitening} refers to a property that when projected onto the whitening metric basis, the resulting preconditioned gradients have an identity covariance:
\begin{equation}
    Cov(\hat{\nabla},\hat{\nabla}) = I \qquad \text{where} \qquad \hat{\nabla} = W^{-1} \nabla_\theta \;L (\theta,x).
\end{equation}
The whitening metric has the same eigenbasis as the empirical Fisher, as they are symmetric matrix powers of each other.

\newpage
\section*{NeurIPS Paper Checklist}

\begin{enumerate}

\item {\bf Claims}
    \item[] Question: Do the main claims made in the abstract and introduction accurately reflect the paper's contributions and scope?
    \item[] Answer: \answerYes{} %
    \item[] Justification: We describe the empirical setting, then show the according results in the paper.
    \item[] Guidelines:
    \begin{itemize}
        \item The answer NA means that the abstract and introduction do not include the claims made in the paper.
        \item The abstract and/or introduction should clearly state the claims made, including the contributions made in the paper and important assumptions and limitations. A No or NA answer to this question will not be perceived well by the reviewers. 
        \item The claims made should match theoretical and experimental results, and reflect how much the results can be expected to generalize to other settings. 
        \item It is fine to include aspirational goals as motivation as long as it is clear that these goals are not attained by the paper. 
    \end{itemize}

\item {\bf Limitations}
    \item[] Question: Does the paper discuss the limitations of the work performed by the authors?
    \item[] Answer: \answerYes{} %
    \item[] Justification: We include a limitations section in the final page.
    \item[] Guidelines:
    \begin{itemize}
        \item The answer NA means that the paper has no limitation while the answer No means that the paper has limitations, but those are not discussed in the paper. 
        \item The authors are encouraged to create a separate "Limitations" section in their paper.
        \item The paper should point out any strong assumptions and how robust the results are to violations of these assumptions (e.g., independence assumptions, noiseless settings, model well-specification, asymptotic approximations only holding locally). The authors should reflect on how these assumptions might be violated in practice and what the implications would be.
        \item The authors should reflect on the scope of the claims made, e.g., if the approach was only tested on a few datasets or with a few runs. In general, empirical results often depend on implicit assumptions, which should be articulated.
        \item The authors should reflect on the factors that influence the performance of the approach. For example, a facial recognition algorithm may perform poorly when image resolution is low or images are taken in low lighting. Or a speech-to-text system might not be used reliably to provide closed captions for online lectures because it fails to handle technical jargon.
        \item The authors should discuss the computational efficiency of the proposed algorithms and how they scale with dataset size.
        \item If applicable, the authors should discuss possible limitations of their approach to address problems of privacy and fairness.
        \item While the authors might fear that complete honesty about limitations might be used by reviewers as grounds for rejection, a worse outcome might be that reviewers discover limitations that aren't acknowledged in the paper. The authors should use their best judgment and recognize that individual actions in favor of transparency play an important role in developing norms that preserve the integrity of the community. Reviewers will be specifically instructed to not penalize honesty concerning limitations.
    \end{itemize}

\item {\bf Theory Assumptions and Proofs}
    \item[] Question: For each theoretical result, does the paper provide the full set of assumptions and a complete (and correct) proof?
    \item[] Answer: \answerNA{} %
    \item[] Justification: We do not provide theoretical results.
    \item[] Guidelines:
    \begin{itemize}
        \item The answer NA means that the paper does not include theoretical results. 
        \item All the theorems, formulas, and proofs in the paper should be numbered and cross-referenced.
        \item All assumptions should be clearly stated or referenced in the statement of any theorems.
        \item The proofs can either appear in the main paper or the supplemental material, but if they appear in the supplemental material, the authors are encouraged to provide a short proof sketch to provide intuition. 
        \item Inversely, any informal proof provided in the core of the paper should be complemented by formal proofs provided in appendix or supplemental material.
        \item Theorems and Lemmas that the proof relies upon should be properly referenced. 
    \end{itemize}

    \item {\bf Experimental Result Reproducibility}
    \item[] Question: Does the paper fully disclose all the information needed to reproduce the main experimental results of the paper to the extent that it affects the main claims and/or conclusions of the paper (regardless of whether the code and data are provided or not)?
    \item[] Answer: \answerYes{} %
    \item[] Justification: We have put forth a best-faith effort to provide reproducible training details and methodology in section 5. Nuanced specifics can be found within the provided code.
    \item[] Guidelines:
    \begin{itemize}
        \item The answer NA means that the paper does not include experiments.
        \item If the paper includes experiments, a No answer to this question will not be perceived well by the reviewers: Making the paper reproducible is important, regardless of whether the code and data are provided or not.
        \item If the contribution is a dataset and/or model, the authors should describe the steps taken to make their results reproducible or verifiable. 
        \item Depending on the contribution, reproducibility can be accomplished in various ways. For example, if the contribution is a novel architecture, describing the architecture fully might suffice, or if the contribution is a specific model and empirical evaluation, it may be necessary to either make it possible for others to replicate the model with the same dataset, or provide access to the model. In general. releasing code and data is often one good way to accomplish this, but reproducibility can also be provided via detailed instructions for how to replicate the results, access to a hosted model (e.g., in the case of a large language model), releasing of a model checkpoint, or other means that are appropriate to the research performed.
        \item While NeurIPS does not require releasing code, the conference does require all submissions to provide some reasonable avenue for reproducibility, which may depend on the nature of the contribution. For example
        \begin{enumerate}
            \item If the contribution is primarily a new algorithm, the paper should make it clear how to reproduce that algorithm.
            \item If the contribution is primarily a new model architecture, the paper should describe the architecture clearly and fully.
            \item If the contribution is a new model (e.g., a large language model), then there should either be a way to access this model for reproducing the results or a way to reproduce the model (e.g., with an open-source dataset or instructions for how to construct the dataset).
            \item We recognize that reproducibility may be tricky in some cases, in which case authors are welcome to describe the particular way they provide for reproducibility. In the case of closed-source models, it may be that access to the model is limited in some way (e.g., to registered users), but it should be possible for other researchers to have some path to reproducing or verifying the results.
        \end{enumerate}
    \end{itemize}

\item {\bf Open access to data and code}
    \item[] Question: Does the paper provide open access to the data and code, with sufficient instructions to faithfully reproduce the main experimental results, as described in supplemental material?
    \item[] Answer: \answerYes{} %
    \item[] Justification: We only utilize open-source datasets in this work, and provide access to our code.
    \item[] Guidelines:
    \begin{itemize}
        \item The answer NA means that paper does not include experiments requiring code.
        \item Please see the NeurIPS code and data submission guidelines (\url{https://nips.cc/public/guides/CodeSubmissionPolicy}) for more details.
        \item While we encourage the release of code and data, we understand that this might not be possible, so “No” is an acceptable answer. Papers cannot be rejected simply for not including code, unless this is central to the contribution (e.g., for a new open-source benchmark).
        \item The instructions should contain the exact command and environment needed to run to reproduce the results. See the NeurIPS code and data submission guidelines (\url{https://nips.cc/public/guides/CodeSubmissionPolicy}) for more details.
        \item The authors should provide instructions on data access and preparation, including how to access the raw data, preprocessed data, intermediate data, and generated data, etc.
        \item The authors should provide scripts to reproduce all experimental results for the new proposed method and baselines. If only a subset of experiments are reproducible, they should state which ones are omitted from the script and why.
        \item At submission time, to preserve anonymity, the authors should release anonymized versions (if applicable).
        \item Providing as much information as possible in supplemental material (appended to the paper) is recommended, but including URLs to data and code is permitted.
    \end{itemize}

\item {\bf Experimental Setting/Details}
    \item[] Question: Does the paper specify all the training and test details (e.g., data splits, hyperparameters, how they were chosen, type of optimizer, etc.) necessary to understand the results?
    \item[] Answer: \answerYes{} %
    \item[] Justification: We have put forth a best-faith effort to provide reproducible training details and methodology in section 5. Nuanced specifics can be found within the provided code.
    \item[] Guidelines:
    \begin{itemize}
        \item The answer NA means that the paper does not include experiments.
        \item The experimental setting should be presented in the core of the paper to a level of detail that is necessary to appreciate the results and make sense of them.
        \item The full details can be provided either with the code, in appendix, or as supplemental material.
    \end{itemize}

\item {\bf Experiment Statistical Significance}
    \item[] Question: Does the paper report error bars suitably and correctly defined or other appropriate information about the statistical significance of the experiments?
    \item[] Answer: \answerNo{} %
    \item[] Justification: Due to computational constraints, we are only able to run one trial each for each optimizer, under a specific objective, checkpoint, and learning rate. That said, we control for random seed and data order, and heuristically the training runs do not large independent variance.
    \item[] Guidelines:
    \begin{itemize}
        \item The answer NA means that the paper does not include experiments.
        \item The authors should answer "Yes" if the results are accompanied by error bars, confidence intervals, or statistical significance tests, at least for the experiments that support the main claims of the paper.
        \item The factors of variability that the error bars are capturing should be clearly stated (for example, train/test split, initialization, random drawing of some parameter, or overall run with given experimental conditions).
        \item The method for calculating the error bars should be explained (closed form formula, call to a library function, bootstrap, etc.)
        \item The assumptions made should be given (e.g., Normally distributed errors).
        \item It should be clear whether the error bar is the standard deviation or the standard error of the mean.
        \item It is OK to report 1-sigma error bars, but one should state it. The authors should preferably report a 2-sigma error bar than state that they have a 96\% CI, if the hypothesis of Normality of errors is not verified.
        \item For asymmetric distributions, the authors should be careful not to show in tables or figures symmetric error bars that would yield results that are out of range (e.g. negative error rates).
        \item If error bars are reported in tables or plots, The authors should explain in the text how they were calculated and reference the corresponding figures or tables in the text.
    \end{itemize}

\item {\bf Experiments Compute Resources}
    \item[] Question: For each experiment, does the paper provide sufficient information on the computer resources (type of compute workers, memory, time of execution) needed to reproduce the experiments?
    \item[] Answer: \answerYes{} %
    \item[] Justification: We describe the machines used to run the experiments, and give an estimate of the time taken.
    \item[] Guidelines:
    \begin{itemize}
        \item The answer NA means that the paper does not include experiments.
        \item The paper should indicate the type of compute workers CPU or GPU, internal cluster, or cloud provider, including relevant memory and storage.
        \item The paper should provide the amount of compute required for each of the individual experimental runs as well as estimate the total compute. 
        \item The paper should disclose whether the full research project required more compute than the experiments reported in the paper (e.g., preliminary or failed experiments that didn't make it into the paper). 
    \end{itemize}
    
\item {\bf Code Of Ethics}
    \item[] Question: Does the research conducted in the paper conform, in every respect, with the NeurIPS Code of Ethics \url{https://neurips.cc/public/EthicsGuidelines}?
    \item[] Answer: \answerYes{} %
    \item[] Justification: We follow the ethics guidelines.
    \item[] Guidelines:
    \begin{itemize}
        \item The answer NA means that the authors have not reviewed the NeurIPS Code of Ethics.
        \item If the authors answer No, they should explain the special circumstances that require a deviation from the Code of Ethics.
        \item The authors should make sure to preserve anonymity (e.g., if there is a special consideration due to laws or regulations in their jurisdiction).
    \end{itemize}

\item {\bf Broader Impacts}
    \item[] Question: Does the paper discuss both potential positive societal impacts and negative societal impacts of the work performed?
    \item[] Answer: \answerNA{} %
    \item[] Justification: We do not find the need to discuss specific societal impacts of this work, as it purely relates to a computational problem.
    \item[] Guidelines:
    \begin{itemize}
        \item The answer NA means that there is no societal impact of the work performed.
        \item If the authors answer NA or No, they should explain why their work has no societal impact or why the paper does not address societal impact.
        \item Examples of negative societal impacts include potential malicious or unintended uses (e.g., disinformation, generating fake profiles, surveillance), fairness considerations (e.g., deployment of technologies that could make decisions that unfairly impact specific groups), privacy considerations, and security considerations.
        \item The conference expects that many papers will be foundational research and not tied to particular applications, let alone deployments. However, if there is a direct path to any negative applications, the authors should point it out. For example, it is legitimate to point out that an improvement in the quality of generative models could be used to generate deepfakes for disinformation. On the other hand, it is not needed to point out that a generic algorithm for optimizing neural networks could enable people to train models that generate Deepfakes faster.
        \item The authors should consider possible harms that could arise when the technology is being used as intended and functioning correctly, harms that could arise when the technology is being used as intended but gives incorrect results, and harms following from (intentional or unintentional) misuse of the technology.
        \item If there are negative societal impacts, the authors could also discuss possible mitigation strategies (e.g., gated release of models, providing defenses in addition to attacks, mechanisms for monitoring misuse, mechanisms to monitor how a system learns from feedback over time, improving the efficiency and accessibility of ML).
    \end{itemize}
    
\item {\bf Safeguards}
    \item[] Question: Does the paper describe safeguards that have been put in place for responsible release of data or models that have a high risk for misuse (e.g., pretrained language models, image generators, or scraped datasets)?
    \item[] Answer: \answerNA{} %
    \item[] Justification: We propose a computational method, and do not release any new data.
    \item[] Guidelines:
    \begin{itemize}
        \item The answer NA means that the paper poses no such risks.
        \item Released models that have a high risk for misuse or dual-use should be released with necessary safeguards to allow for controlled use of the model, for example by requiring that users adhere to usage guidelines or restrictions to access the model or implementing safety filters. 
        \item Datasets that have been scraped from the Internet could pose safety risks. The authors should describe how they avoided releasing unsafe images.
        \item We recognize that providing effective safeguards is challenging, and many papers do not require this, but we encourage authors to take this into account and make a best faith effort.
    \end{itemize}

\item {\bf Licenses for existing assets}
    \item[] Question: Are the creators or original owners of assets (e.g., code, data, models), used in the paper, properly credited and are the license and terms of use explicitly mentioned and properly respected?
    \item[] Answer: \answerYes{} %
    \item[] Justification: We cite the two open-source datasets that we utilize.
    \item[] Guidelines:
    \begin{itemize}
        \item The answer NA means that the paper does not use existing assets.
        \item The authors should cite the original paper that produced the code package or dataset.
        \item The authors should state which version of the asset is used and, if possible, include a URL.
        \item The name of the license (e.g., CC-BY 4.0) should be included for each asset.
        \item For scraped data from a particular source (e.g., website), the copyright and terms of service of that source should be provided.
        \item If assets are released, the license, copyright information, and terms of use in the package should be provided. For popular datasets, \url{paperswithcode.com/datasets} has curated licenses for some datasets. Their licensing guide can help determine the license of a dataset.
        \item For existing datasets that are re-packaged, both the original license and the license of the derived asset (if it has changed) should be provided.
        \item If this information is not available online, the authors are encouraged to reach out to the asset's creators.
    \end{itemize}

\item {\bf New Assets}
    \item[] Question: Are new assets introduced in the paper well documented and is the documentation provided alongside the assets?
    \item[] Answer: \answerNA{} %
    \item[] Justification: We do not release new assets.
    \item[] Guidelines:
    \begin{itemize}
        \item The answer NA means that the paper does not release new assets.
        \item Researchers should communicate the details of the dataset/code/model as part of their submissions via structured templates. This includes details about training, license, limitations, etc. 
        \item The paper should discuss whether and how consent was obtained from people whose asset is used.
        \item At submission time, remember to anonymize your assets (if applicable). You can either create an anonymized URL or include an anonymized zip file.
    \end{itemize}

\item {\bf Crowdsourcing and Research with Human Subjects}
    \item[] Question: For crowdsourcing experiments and research with human subjects, does the paper include the full text of instructions given to participants and screenshots, if applicable, as well as details about compensation (if any)? 
    \item[] Answer: \answerNA{} %
    \item[] Justification: We do not involve crowdsourcing or research with human subjects.
    \item[] Guidelines:
    \begin{itemize}
        \item The answer NA means that the paper does not involve crowdsourcing nor research with human subjects.
        \item Including this information in the supplemental material is fine, but if the main contribution of the paper involves human subjects, then as much detail as possible should be included in the main paper. 
        \item According to the NeurIPS Code of Ethics, workers involved in data collection, curation, or other labor should be paid at least the minimum wage in the country of the data collector. 
    \end{itemize}

\item {\bf Institutional Review Board (IRB) Approvals or Equivalent for Research with Human Subjects}
    \item[] Question: Does the paper describe potential risks incurred by study participants, whether such risks were disclosed to the subjects, and whether Institutional Review Board (IRB) approvals (or an equivalent approval/review based on the requirements of your country or institution) were obtained?
    \item[] Answer: \answerNA{} %
    \item[] Justification: We do not involve crowdsourcing or research with human subjects.
    \item[] Guidelines:
    \begin{itemize}
        \item The answer NA means that the paper does not involve crowdsourcing nor research with human subjects.
        \item Depending on the country in which research is conducted, IRB approval (or equivalent) may be required for any human subjects research. If you obtained IRB approval, you should clearly state this in the paper. 
        \item We recognize that the procedures for this may vary significantly between institutions and locations, and we expect authors to adhere to the NeurIPS Code of Ethics and the guidelines for their institution. 
        \item For initial submissions, do not include any information that would break anonymity (if applicable), such as the institution conducting the review.
    \end{itemize}

\end{enumerate}

\end{document}